\documentclass[sigconf]{acmart}

\AtBeginDocument{%
  }

\usepackage{algorithm}
\usepackage{enumitem}
\usepackage{multirow}
\usepackage{pifont}

\usepackage{algorithmic}

\usepackage{amsmath,amsfonts,bm}

\def\eqref#1{equation~\ref{#1}}

\def\1{\bm{1}}

\def\vv{{\bm{v}}}

\def\vy{{\bm{y}}}

\DeclareMathAlphabet{\mathsfit}{\encodingdefault}{\sfdefault}{m}{sl}
\SetMathAlphabet{\mathsfit}{bold}{\encodingdefault}{\sfdefault}{bx}{n}

\def\gD{{\mathcal{D}}}

\def\gG{{\mathcal{G}}}

\def\gL{{\mathcal{L}}}
\def\gM{{\mathcal{M}}}

\def\gS{{\mathcal{S}}}

\def\gV{{\mathcal{V}}}

\theoremstyle{plain}
\newtheorem{theorem}{Theorem}[section]

\theoremstyle{definition}

\newtheorem{assumption}[theorem]{Assumption}
\theoremstyle{remark}

\newcommand{\model}{\textsc{NodeImport}{}}

\newcommand{\ms}[2]{{#1\tiny{$\pm$#2}}}

\definecolor{d3blue}{HTML}{1f77b4}
\definecolor{d3orange}{HTML}{ff7f0e}
\definecolor{d3green}{HTML}{2ca02c}
\definecolor{d3red}{HTML}{d62728}

\newcommand{\cmark}{\ding{51}}%
\newcommand*\colourcheck[1]{%
  \expandafter\newcommand\csname #1check\endcsname{\textcolor{#1}{\ding{51}}}%
}
\newcommand*\colourx[1]{%
  \expandafter\newcommand\csname #1x\endcsname{\textcolor{#1}{\ding{55}}}%
}
\colourcheck{blue}
\colourcheck{green}
\colourx{gray}

\newcommand{\ie}{{\it i.e.}}
\newcommand{\stitle}[1]{\vspace{0.8mm} \noindent {\bf #1}}

\newcommand\Nan[1]{{#1}}

\copyrightyear{2025}
\acmYear{2025}
\setcopyright{rightsretained}
\acmConference[KDD '25]{Proceedings of the 31st ACM SIGKDD Conference on Knowledge Discovery and Data Mining V.1}{August 3--7, 2025}{Toronto, ON, Canada}
\acmBooktitle{Proceedings of the 31st ACM SIGKDD Conference on Knowledge Discovery and Data Mining V.1 (KDD '25), August 3--7, 2025, Toronto, ON, Canada}
\acmDOI{10.1145/3690624.3709215}
\acmISBN{979-8-4007-1245-6/25/08}

\begin{document}


\title{\model: Imbalanced Node Classification with \\ Node Importance Assessment}

\author{Nan Chen}
\affiliation{%
  \institution{Johns Hopkins University}
  \city{Baltimore}
  \state{Maryland}
  \country{USA}
}
\email{nchen38@jh.edu}
\orcid{0009-0007-0240-5748}

\author{Zemin Liu$^*$}
\affiliation{%
  \institution{Zhejiang University}
  \city{Hangzhou}
  \state{Zhejiang}
  \country{China}}
\email{liu.zemin@zju.edu.cn}
\orcid{0000-0001-6262-9435}

\author{Bryan Hooi}
\affiliation{%
  \institution{National University of Singapore}
  \city{Singapore}
  \country{Singapore}
}
\email{bhooi@comp.nus.edu.sg}
\orcid{0000-0002-5645-1754}

\author{Bingsheng He$^*$}
\affiliation{%
 \institution{National University of Singapore}
 \city{Singapore}
 \country{Singapore}}
 \email{hebs@comp.nus.edu.sg}
 \orcid{0000-0001-8618-4581}

\author{Jun Hu}
\affiliation{%
  \institution{National University of Singapore}
  \city{Singapore}
  \country{Singapore}}
  \email{jun.hu@nus.edu.sg}
  \orcid{0000-0003-1277-6802}

\author{Jia Chen}
\affiliation{%
  \institution{Grabtaxi Holdings Pte Ltd}
  \city{Singapore}
  \country{Singapore}}
\email{jia.chen@grab.com}
\orcid{0009-0003-1174-0063}

\thanks{
    $^*$The corresponding authors.
}

\renewcommand{\shortauthors}{Nan Chen et al.}

\begin{abstract}
In real-world applications, node classification on graphs often faces the challenge of class imbalance, where majority classes dominate training, resulting in biased model performance. Traditional Graph Neural Networks (GNNs) often struggle in such scenarios, as they tend to overfit to majority classes while underrepresenting minority classes. Existing solutions, which either prioritize nodes based on class size or synthesize new nodes for minority classes, often fall short of effectively addressing this imbalance issue. This paper introduces a novel approach to class-imbalanced node classification by utilizing a balanced meta-set for importance measurement, where a training node is considered significant if it enhances model performance under an unbiased setting. Our method identifies important nodes that can counteract class imbalance and utilizes them for model training, allowing for fine-grained and dynamic node selection throughout the training process. We theoretically derive a formula to directly assess node importance, reducing computational overhead and providing an intuitive threshold for node selection. Guided by this metric, we develop a novel framework that filters valuable labeled, unlabeled, and synthetic nodes that enhance model performance in an unbiased context. A key advantage of this framework is its separation of the synthetic node generation process from the filtering process, ensuring compatibility with various node generation techniques. Furthermore, we introduce a strategy to construct a high-quality meta-set that closely approximates the overall feature distribution, ensuring robust representation of each class. We evaluate our framework, {\textsc{NodeImport}}, across multiple benchmark datasets using popular GNN architectures, demonstrating its superiority over state-of-the-art baselines. Our results highlight the flexibility and effectiveness of the framework in mitigating class imbalance, leading to improved node classification outcomes. The source code is available at \url{https://github.com/NanChanNN/NodeImport}.
\end{abstract}




\maketitle

\section{Introduction}
Graph data is prevalent across many domains, making graph analysis, particularly node classification, a significant research focus \citep{cai2018comprehensive,wu2020comprehensive}. With the rise of deep learning, Graph Neural Networks (GNNs) \citep{kipf2017semisupervised,hamilton2017inductive,velivckovic2017graph,xu2018powerful} have become the go-to method for node-level graph analysis, achieving notable success on benchmark datasets. However, GNNs often falter when applied to real-world graphs with imbalanced node class distributions \citep{zhao2021graphsmote,zhuo2024partitioning,song2022tam,chen2024consistency,Zhou_Gong_2023graphsr}. In these scenarios, majority classes have significantly more labeled nodes for training than minority classes, leading to performance bias that favors majority classes and marginalizes minority classes, resulting in sub-optimal classification outcomes.

To address class-imbalanced node classification, researchers have developed various solutions, which can be broadly categorized into algorithm-level and data-level approaches \citep{liu2023survey}. Most of these methods can be regarded as identifying and prioritizing important nodes during training to mitigate the imbalance issue. Algorithm-level approaches enhance the importance of minority classes by assigning higher weights to them or expanding their margins \citep{japkowicz2002class,ren2020balanced,hong2021disentangling,chen2021topology,song2022tam}. However, these methods typically determine a node's importance based solely on its class size, disregarding its specific features or position within the graph. This oversight of intrinsic differences among nodes of the same class oversimplifies the problem and poses challenges to model training. For instance, outliers within minority classes may introduce noise, and overemphasizing these outliers can skew the model towards misleading patterns. Additionally, certain instances within majority classes, especially those near class boundaries, may be crucial in forming a clear decision boundary that benefits both majority and minority classes. Furthermore, static weighting mechanisms are agnostic to the model's current status and cannot adapt to the model's training progress. Since neural networks tend to learn patterns of increasing complexity as training progresses \citep{yang2023towards}, a dynamic weighting mechanism that adapts to the model's learning progress is preferable to a fixed scheme.

On the other hand, data-level approaches balance class sizes by oversampling minority classes through node synthesis \cite{zhao2021graphsmote,wu2021graphmixup,park2022graphens,li2023graphsha,liu2023imbalanced}. Most methods follow a MixUp-like style \citep{Zhang2019mixup,verma2019manifold}, sampling pairs of nodes and mixing their features and neighbors to generate new nodes. The generation process serves as an implicit measure of importance, with various heuristics refining each stage of the synthesis pipeline to produce potentially important nodes. However, data-level approaches have limited flexibility, as their implicit importance measures cannot be easily extended to unlabeled nodes, which have significant potential to augment the dataset and address the class imbalance issue~\cite{berthelot2019mixmatch,sohn2020fixmatch,feng2020grand,Zhou_Gong_2023graphsr}.

In this work, we propose a different strategy to identify valuable nodes for training to address class-imbalanced node classification, using a balanced meta-set for importance measurement. Our approach is based on a simple yet fundamental assumption \cite{ren2018learning,shu2019meta,ren2020balanced,mindermann2022prioritized}: a training node is deemed important if the model, after being trained on it, performs better under an unbiased setting. This strategy offers two main advantages. First, by evaluating the contributions of individual nodes to model performance in a balanced scenario, we can dynamically distinguish important nodes from negligible ones within a class at different training phases. Second, this assumption allows for increased flexibility. For instance, with node oversampling, we can utilize any off-the-shelf node synthesis techniques to generate new nodes and then use the balanced meta-set to identify those that can alleviate class imbalance.

However, straightforwardly applying a meta-set for node filtering in graph-structured data poses unique challenges, leading to two urgent issues. First, assessing node importance with a meta-set requires a bi-level evaluation process: training the model on each node individually and then evaluating the trained model on the meta-set. This process is computationally demanding and time-consuming, which calls for a relaxation of the bi-level evaluation process to achieve higher efficiency. Second, in class-imbalanced node classification scenarios, the balanced meta-set we could construct is generally quite small, making its quality sensitive to random noise. Since the quality of the meta-set is crucial to the overall performance, a carefully designed method is needed to build a high-quality meta-set from the training set.

To tackle these issues, we introduce {\model}, a versatile framework for class-imbalanced node classification tasks that employs an unbiased meta-set to identify important nodes for training. By expanding the bi-level evaluation process and adopting a series of graph-specific assumptions, we derive a formula to directly calculate the importance of a node in the graph, thereby reducing computational overhead. The derived formula indicates that node importance is influenced not only by static graph characteristics but also by dynamic model prediction behaviors, providing an intuitive interpretation of the formula. Moreover, this formula naturally gives rise to a meaningful threshold for node selection, eliminating the need for manual threshold tuning. Leveraging our importance formula, we can seamlessly incorporate unlabeled and synthetic nodes into model training, in addition to labeled nodes. Since the computation of node importance is orthogonal to the node generation process, we decouple the node generation process from the filtering process, allowing us to use any existing methods to synthesize nodes or create pseudo-labels for unlabeled nodes, while the filtering process, using the importance formula, then selects valuable nodes able to address class imbalance. Furthermore, to build a high-quality meta-set, we conduct clustering within the embedding space of each class and select representative samples as meta-samples to capture the intrinsic patterns of each class.

We perform a comprehensive empirical evaluation of {\model} using three popular GNN architectures: GCN \citep{kipf2017semisupervised}, GAT \citep{velivckovic2017graph}, and GraphSAGE \citep{hamilton2017inductive}. This evaluation utilizes multiple benchmark datasets, comprising three citation datasets \citep{yang2016revisiting} and two Amazon co-purchase datasets \citep{shchur2018pitfalls}, under a long-tailed imbalance setting \citep{cui2019class}. The experimental results highlight the effectiveness of the derived importance formula for selecting valuable nodes for training. In addition, it demonstrates the efficiency of the proposed framework in incorporating various node sources to augment the training set and address class imbalance in node-level classification tasks. Overall, the major contributions of this work are summarized as:
\begin{itemize}
    \item To address class imbalance in node classification, we propose using a balanced meta-set to select nodes valuable to model training. This approach allows a detailed examination of individual nodes' contributions and provides increased flexibility in the node selection process.
    \item We theoretically derive a formula to directly assess node importance, thus largely reducing the computational overhead. Furthermore, this formula naturally gives rise to a filtering threshold and has an intuitive interpretation.
    \item Leveraging the derived formula, we introduce {\model}, a versatile framework that incorporates valuable labeled, unlabeled, and synthetic nodes for training to counter class distribution imbalance. Additionally, we build a component to extract a high-quality meta-set from labeled nodes.
    \item Comprehensive experiments on various public benchmark datasets demonstrate the superiority of {\model} compared to state-of-the-art baselines. Detailed analyses highlight the necessity and effectiveness of its components.
\end{itemize}

\section{Related works}

\stitle{Class imbalance classification.}
Existing solutions to the class imbalance problem can be broadly categorized into algorithm-level and data-level approaches~\citep{liu2023survey}. Algorithm-level methods aim to guide the model to focus more on minority classes. This can be achieved by assigning higher weights to minority classes~\citep{japkowicz2002class,lin2017focal,cui2019class,xu2020class,GuoWZL24} or by expanding the margins for minority classes~\citep{cao2019learning,ren2020balanced,tan2020equalization,menon2021longtail,hong2021disentangling}. However, these methods gauge a training sample's importance according to its class size, treating all samples from minority classes as equally important without considering their intrinsic characteristics. Although some methods tailored for graph data, such as TAM~\citep{song2022tam} and ReNode~\citep{chen2021topology}, consider a node's topological location when assigning weights, their weights do not dynamically adapt to the model's status during training. Other algorithm-level approaches include imposing additional regularization constraints~\citep{drgcn,yan2023rethinking}, employing multi-expert training with knowledge distillation~\citep{yun2022lte4g}, or \Nan{by correcting nodes with ineffective message passing caused by topological disparities.~\citep{Liu24BAT}}.

 Data-level techniques balance class sizes through resampling~\citep{liu2021pick,zhang2021bag,yun2022lte4g,cui2022allie} or generative methods~\citep{chawla2002smote,qu2021imgagn,wang2021rsg}. In the graph domain, most works~\citep{zhao2021graphsmote,wu2021graphmixup,park2022graphens,li2023graphsha,liu2023imbalanced} adopt MixUp-like~\citep{Zhang2019mixup,verma2019manifold} generative methods to augment minority classes, differing in how they sample node pairs, mix node features, and construct edges for synthetic nodes. These generative techniques synthesize potentially important minority nodes based on heuristics adopted in the synthesis pipeline. However, these heuristics implicitly define an importance measure that cannot be directly applied to unlabeled nodes, thus limiting their flexibility. Recently, GraphSR~\citep{Zhou_Gong_2023graphsr} was proposed to identify valuable unlabeled nodes to augment minority classes, but its rules cannot be extended to evaluate synthetic nodes. In contrast, our work employs a balanced meta-set to identify important nodes capable of handling class imbalance for training, positing that a training node is essential if it can improve the model's performance under an unbiased condition.

For an in-depth discussion of class imbalance learning, please kindly refer to the following surveys~\citep{he2009learning,krawczyk2016learning,ma2023class,zhang2023deep,liu2023survey}.

\stitle{Meta-set guided data selection.}
In the image domain, several works employ a balanced meta-set to learn instance-wise weights~\citep{ren2018learning,shu2019meta} or class-wise sampling rates~\citep{ren2020balanced} to counter class imbalance. However, these approaches often require bi-level optimization during training, imposing a significant computational burden. On the contrary, we focus on graph-structured data and theoretically derive a formula to simplify the bi-level optimization process, thus saving computing resources. In addition, the derived formula naturally gives a filtering threshold and has a specific interpretation related to the graph domain.

In the active learning domain, meta-sets are used to identify valuable training nodes~\citep{kirsch2021test,killamsetty2021glister,mindermann2022prioritized}, aiming to train a model with as few samples as possible to speed up the training process. Meta-sets in these works usually bear the same class distribution as the training set and do not need to be balanced. The most relevant work to our derived formula is the RHO-Loss~\citep{mindermann2022prioritized}, which derives a formula to simplify the computation of sample importance. However, the RHO-Loss still requires an extra model trained on the meta-set and a pre-defined filtering threshold. By contrast, our work customizes the RHO-Loss to the graph domain, deriving a formula with semantic meaning specific to graphs. This formula is more straightforward and does not require storing an additional model trained on the meta-set, thus cutting down memory consumption. Additionally, it naturally provides a filtering threshold for node selection, eliminating the need for manual threshold tuning.

\section{Preliminary}

\stitle{Graph representation learning.}
A graph \( G \) is represented as \( \mathcal{G} = \{\mathcal{V}, \mathcal{E}, X\} \), where \( \mathcal{V} \) is the set of \( n \) nodes, \( \mathcal{E} \subseteq \mathcal{V} \times \mathcal{V} \) is the set of edges, and \( X \in \mathbb{R}^{n \times d} \) is the node feature matrix with each row \( x_v \in \mathbb{R}^d \) being the feature vector for node \( v \in \mathcal{V} \). The graph structure can also be described by an adjacency matrix \( A \in \{0,1\}^{n \times n} \), where \( A_{i,j} = 1 \) if there is an edge \( \langle i,j \rangle \) and \( A_{i,j} = 0 \) otherwise. Each node \( v \) is associated with a one-hot encoded label vector \( y_v \in \{0,1\}^{1 \times c} \) over \( c \) classes, and \( Y \in \{0,1\}^{n \times c} \) is the label matrix containing these label vectors. Given a graph encoder \( g(\cdot; \theta) \) with parameters \( \theta \) (e.g., a GNN model), nodes within the graph can be mapped into probability distributions \( H \in \mathbb{R}^{n \times c} \) indicating the likelihood of each node belonging to different classes, formulated as \( H = g(\mathcal{G}; \theta) \). We denote the predictive distribution of the GNN model on node \( v \) as \( p(y_v | v; \theta) \) for the ease of analysis.

\stitle{Class-imbalanced node classification.}
Class-imbalanced node classification involves categorizing nodes into their respective classes when the label distribution among classes is uneven, often resulting in a performance bias favoring majority classes \citep{shi2020multi,zhao2021graphsmote}. The imbalance ratio $IR=r_0/r_1$ quantifies the degree of imbalance, where $r_0$ and $r_1$ represent the sizes of the largest and smallest classes, respectively \citep{liu2023survey}. The optimal parameters of the graph encoder $\theta^*$ are obtained by minimizing the cross-entropy loss on the training set $\gD_{tr}=\{(v, y_v)\}$ defined as:
\begin{equation} \label{eq.main-loss}
 \textstyle   \theta^* = \underset{\theta}{\arg\min} \sum_{(v,y_v)\in \gD_{tr}} - \text{log } p(y_v | v ; \theta).
\end{equation}
Here, we utilize $L[y_v | v; \theta]=-\text{log } p(y_v | v ; \theta)$ to denote the cross-entropy loss for an individual training point $(v, y_v)\in\gD_{tr}$.

\stitle{Importance assessment using meta-sets.}
We introduce an unbiased meta-set $\gD_{meta}=\{(v^{(meta)}, y_v^{(meta)})\}$ with a balanced class distribution for evaluating node importance. For simplicity, this meta-set is written as $\vv^{(meta)}$ and $\vy^{(meta)}$. The meta-set is usually much smaller than the training set (i.e., $|\gD_{tr}| \gg |\gD_{meta}|$), and helps rectify the graph encoder in cases of imbalanced training data. Given a training point $(v, y_v) \in \gD_{tr}$, we define the following metric $\eta_v$ to evaluate its importance to model training:
\begin{equation} \label{eq.model-objective}
\begin{aligned}
  \eta_v &= \text{log} \frac{p(\vy^{(meta)}|\vv^{(meta)}; \theta_t; (v,y_v))}{ p(\vy^{(meta)}|\vv^{(meta)}; \theta_t)} \\ 
  &= \text{log } p(\vy^{(meta)}|\vv^{(meta)}; \theta_t; (v,y_v)) \\
  &\quad - \text{log } p(\vy^{(meta)}|\vv^{(meta)}; \theta_t). 
\end{aligned}
\end{equation}
Intuitively, this metric assesses whether the current model's performance on the balanced meta-set improves after being trained on the given point. When $\eta_v > 0$, the node is considered important and used in the next training step. Calculating this metric involves training the current model on each data point individually and determining if the updated model performs better on the meta-set. This process is computationally intensive, which necessitates a more tractable approximation.

\section{The Proposed Model: \model}

\begin{figure*}[!ht]
  \centering
  \includegraphics[width=0.99\linewidth]{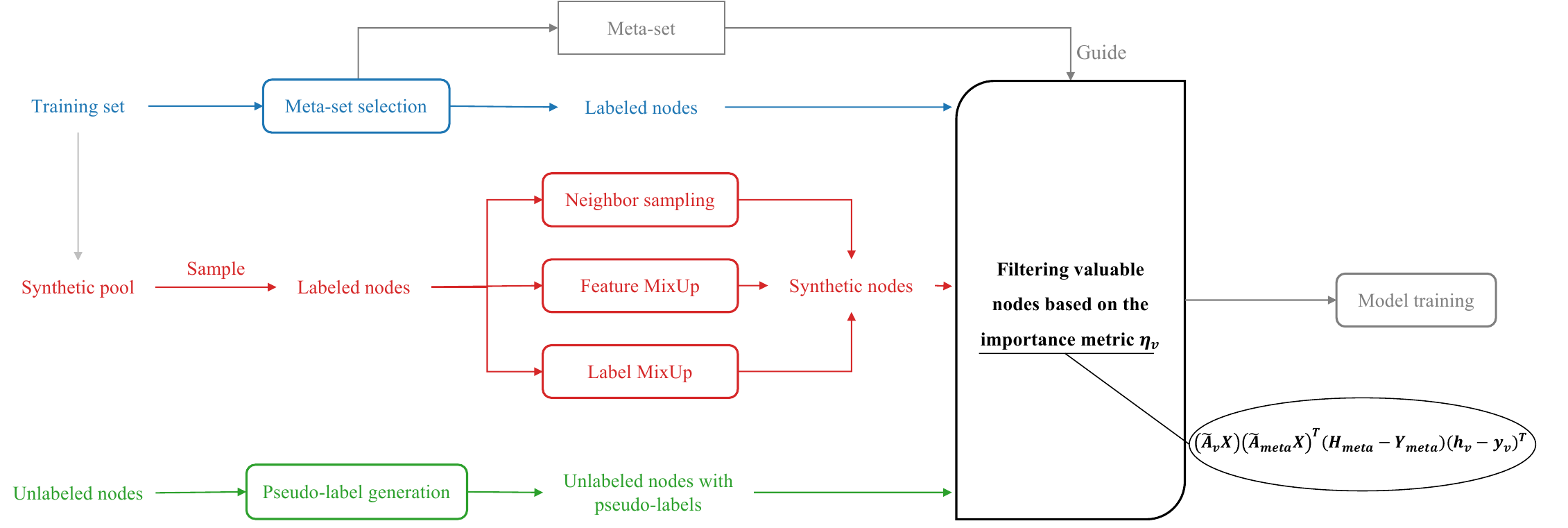}
  \caption{The structure of the proposed framework. The 	\textcolor{d3blue}{blue}, \textcolor{d3green}{green}, and \textcolor{d3red}{red} sections indicate component \textcolor{d3blue}{1}, \textcolor{d3green}{2}, and \textcolor{d3red}{3}, respectively.}
  \Description{Framework structure.}
  \label{figure:framework_architectur}
\end{figure*}

In this section, we start by deriving a tractable formula to approximate Eq.~\ref{eq.model-objective}, which simplifies the bi-level selection process to pinpoint important nodes. Subsequently, we illustrate how this formula can be utilized to identify crucial labeled, unlabeled, and synthetic nodes to augment the training set. Finally, we introduce our meta-set construction method, which selects representative meta-samples. The entire framework architecture is illustrated in Figure~\ref{figure:framework_architectur}, while its pseudocode and complexity analysis are provided in Appendix~\ref{desc-algorithm}.

\subsection{Tractable Importance Metric for Node Selection}
Following \cite{mindermann2022prioritized}, we employ Bayesian probability theory to derive a more tractable form of Eq.~\ref{eq.model-objective}. By applying Bayes' theorem and assuming conditional independence, $p(y_v|v,u;\theta_t)=p(y_v|v;\theta_t) $, we arrive at the following equation:
\begin{equation} \label{eq:bayes_rule}
\begin{aligned}
    \eta_v &= \text{log } p(\vy^{(meta)}|\vv^{(meta)}; \theta_t; (v,y_v)) \\
    &\quad - \text{log } p(\vy^{(meta)}|\vv^{(meta)}; \theta_t) \\
    &= \text{log } \frac{p(y_v | v; \vv^{(meta)}, \vy^{(meta)}; \theta_t) \cdot p(\vy^{(meta)} | \vv^{(meta)}, v; \theta_t)}{p(y_v|v, \vv^{(meta)}; \theta_t)} \\ 
    &\quad - \text{log } p(\vy^{(meta)}|\vv^{(meta)}; \theta_t) \\
    &= \text{log } \frac{p(y_v | v; \vv^{(meta)}, \vy^{(meta)}; \theta_t) \cdot p(\vy^{(meta)} | \vv^{(meta)}; \theta_t)}{p(y_v|v; \theta_t)} \\
    &\quad - \text{log } p(\vy^{(meta)}|\vv^{(meta)}; \theta_t) \\
    &= \text{log } p(y_v | v; (\vv^{(meta)}, \vy^{(meta)}); \theta_t) - \text{log } p(y_v|v; \theta_t),
\end{aligned}
\end{equation}
where the second equation applies Bayes' theorem to decompose $p(\vy^{(meta)}|\vv^{(meta)}; \theta_t; (v,y_v))$, and the third equation simplifies terms using the assumption of conditional independence. Intuitively, the final equation in Eq.~\ref{eq:bayes_rule} can be interpreted as the difference in performance at the training point $(v,y_v)$ between the current model and the one updated with the meta-set. Let $\hat{\theta}_t$ denote the parameters of the updated model, we have: 
\begin{equation} \label{eq:expression_1}
\begin{aligned}
    \eta_v &= \text{log } p(y_v | v; \hat{\theta}_t) - \text{log } p(y_v|v; \theta_t) \\
    &= L[y_v|v;\theta_t] - L[y_v|v;\hat{\theta}_t],
\end{aligned}
\end{equation}
where, in the second equation, we adopt the definition of cross-entropy loss on point $(v, y_v)$. Given Eq.~\ref{eq:expression_1}, the calculation process simplifies to training the current model on the meta-set and monitoring the change of loss values on training points, which incurs only one extra round of training and is thus more efficient. However, it remains computationally demanding as we need to update the model on the meta-set at each training step.

Suppose that $\hat{\theta}_t = \theta_t + \Delta$, where $\Delta$ represents the parameter updates based on the meta-set, we employ a first-order Taylor series expansion to express $L[y_v|v;\hat{\theta}_t]$ in terms of $L[y_v|v;\theta_t]$ and turn Eq.~\ref{eq:expression_1} into the following format:
\begin{equation}\label{eq:expression_2}
\begin{aligned}
\eta_{v} 
&= L[y_v|v;\theta_t] - L[y_v|v;\theta_t+\Delta] \\
&= L[y_v|v;\theta_t] - (L[y_v|v;\theta_t] + \langle \nabla_\theta L[y_v|v;\theta_t], \Delta \rangle_F + o(\Delta^2)) \\ 
&\approx - \langle \nabla_\theta L[y_v|v;\theta_t], \Delta \rangle_F,
\end{aligned}
\end{equation}
where $\nabla_\theta L[y_v|v;\theta_t]$ is the gradient of the individual loss function with regard to weights $\theta$ at $\theta_t$, and $\langle  \cdot,\cdot \rangle _F$ represents the Frobenius inner product. To get a cheaper estimate of parameter updates $\Delta$, we assume that a single optimization step is taken for model $\theta_t$ on the meta-set $\gD_{meta}$, which gives the following assumption.
\begin{assumption}
    We update the current model weights $\theta_t$ on the meta-set $\gD_{meta}$ through one gradient descent step to obtain $\hat{\theta}_t$, meaning that $\hat{\theta}_t=\theta_t - \kappa \cdot \nabla_\theta L[\vy^{(meta)}|\vv^{(meta)};\theta_t]$ where $\kappa$ is a pre-defined learning rate. Therefore, $\Delta=- \kappa \cdot \nabla_\theta L[\vy^{(meta)}|\vv^{(meta)};\theta_t]$.
\end{assumption}
With the assumption above, the importance of a node is determined by the alignment of gradient descent directions between the training point and the meta-set, which is compatible with findings from previous works on meta-learning~\citep{finn2017model,eshratifar2018gradient,ren2018learning,shu2019meta}:
\begin{equation}\label{eq:expression_3}
\begin{aligned}
\eta_{v} &\approx \kappa \cdot \langle \nabla_\theta L[y_v|v;\theta_t], \nabla_\theta L[\vy^{(meta)}|\vv^{(meta)};\theta_t] \rangle_F.
\end{aligned}
\end{equation}
To obtain a more explicit expression of the cross-entropy loss function $L[y_v|v;\theta_t]$, we make the following assumption about the architecture of the GNN encoders used for ease of analysis.

\begin{assumption}
We assume that the GNN encoder in use follows the format $H=\textsc{SoftMax}(\tilde{A}X\theta)$, 
where $\tilde{A}\in\mathbb{R}^{n\times n}$ is the aggregation matrix, $\theta \in \mathbb{R}^{d\times c}$ the model parameters, and $H\in\mathbb{R}^{n\times c}$ the output of the model. Note that the aggregation matrix $\tilde{A}$ is not necessarily equivalent to the adjacency matrix $A$; it can be any polynomial of $A$, such as $\sum_{k=0}^{\infty}\gamma_k A^k$ in GPR-GNN \citep{chien2020adaptive}, $({D}^{-\frac{1}{2}}{A}{D}^{-\frac{1}{2}})^K$ in SGC \citep{wu2019simplifying}, or $\frac{1}{K} \sum_{k=1}^{K} ((1-\alpha) ({D}^{-\frac{1}{2}}{A}{D}^{-\frac{1}{2}})^k + \alpha I)$ in SSGC \citep{zhu2020simple}.
\end{assumption}
Given this assumption, the individual loss function $L[y_v|v;\theta_t]$ can be expressed in terms of the underlying GNN model. Let $	\tilde{A}_v\in\mathbb{R}^{1\times n}$ be the aggregation matrix associated with node $v$, we get:
\begin{equation} \label{eq:loss_format}
\begin{aligned}
    L[y_v|v;\theta_t]
    &= - \textsc{LogSoftMax}(\tilde{A}_v X \theta)y_v^T,
\end{aligned}
\end{equation}
where $\textsc{LogSoftMax}$ represents the function that applies element-wise logarithm to the output of softmax. To proceed with the derivation of Eq.~\ref{eq:expression_3}, it is necessary to compute the derivative of the cross-entropy loss function $L[\cdot]$ defined in Eq.~\ref{eq:loss_format} with respect to the model parameters $\theta$. Given $H=\textsc{SoftMax}(\tilde{A}X\theta)$, the derivative of the cross-entropy loss with regard to model parameters $\theta$ is:
\begin{equation} \label{eq:matrix_gradient}
\begin{aligned}
    \nabla_\theta L &= (\tilde{A} X)^T(H-Y).
\end{aligned}
\end{equation}
The detailed derivation of Eq.~\ref{eq:matrix_gradient} is provided in Appendix~\ref{app.derivative_derivation}. In essence, we work in differential form to compute the derivative. Substituting Eq.~\ref{eq:matrix_gradient} into Eq.~\ref{eq:expression_3} gives our final importance metric:
\begin{equation}\label{eq:final_expression}
\begin{aligned}
\eta_{v} &\approx \kappa \cdot \langle \nabla_\theta L[y_v|v;\theta_t], \nabla_\theta L[\vy^{(meta)}|\vv^{(meta)};\theta_t] \rangle_F \\
&= \kappa \cdot \text{tr}( (h_v-y_v)^T (\tilde{A}_v X) (\tilde{A}_{meta}X)^T (H_{meta} - Y_{meta}) ) \\
&= \kappa \cdot (\tilde{A}_v X) (\tilde{A}_{meta}X)^T (H_{meta} - Y_{meta}) (h_v-y_v)^T,
\end{aligned}
\end{equation}
where $\tilde{A}_{meta}\in\mathbb{R}^{m\times n}$, $H_{meta}\in \mathbb{R}^{m\times c}$, and $Y_{meta}\in [0,1]^{m\times c}$ represent the aggregation matrix, model prediction, and label matrix associated with nodes in the meta-set, respectively. 

\stitle{Understanding the importance metric} We offer an explanation of the functioning of our derived importance metric in Eq.~\ref{eq:final_expression} by breaking it down into two constituent components: 
\begin{itemize}[left=0.1pt]
\item \textbf{Context similarity }$(\tilde{A}_v X) (\tilde{A}_{meta}X)^T$: This component measures the similarity between the node sample $(v,y_v)$ and the meta-set $\gD_{meta}$ by considering both the input features of the nodes and the structure of their local subgraphs. It captures the local context similarity within the graph, reflecting static graph characteristics and being agnostic to the underlying GNN model.
\item \textbf{Prediction behavior similarity }$(H_{meta} - Y_{meta}) (h_v-y_v)^T$: This component evaluates the similarity in the prediction behavior of the model between the node sample $(v, y_v)$ and the meta-set $\gD_{meta}$. It is relevant to the underlying GNN model and captures the model's dynamics during training.
\end{itemize}
By examining the importance metric as a whole, we observe that training samples from under-represented classes (i.e., minority classes) are more favored by the metric compared to those from over-represented classes (i.e., majority classes). This is partly because our meta-set is balanced, ensuring that each class has an equal voting opportunity. Additionally, the term $(H_{meta}-Y_{meta})$ specifically assesses the current model performance on each class and places more emphasis on classes with poor performance. Furthermore, the context similarity and prediction behavior similarity between a node and the meta-samples of its class are also crucial, aiding in the removal of outliers in the minority classes. This explains why our metric is capable of effectively addressing class-imbalanced issues.

\stitle{Instantiating the importance metric.} In practice, we employ existing
polynomials of the adjacency matrix $A$ (such as those used in APPNP~\citep{gasteiger2018predict} or SSGC~\citep{zhu2020simple}) to calculate the aggregation matrix $\tilde{A}$ and utilize the calculated matrix to compute the context similarity. It is important to note that the context similarity is computed once at the beginning of the program and is used throughout the entire training process. The prediction behavior similarity is computed using the output of the current model, which requires no additional computation. In general, applying our formula to evaluate the importance of nodes incurs minimal extra computational overhead.

\subsection{Framework} \label{sec:framework_introduction}
Based on the importance metric derived in Eq.\ref{eq:final_expression}, our primary approach is to apply this metric to labeled nodes. This allows us to filter out valuable nodes for model training, which can effectively counter class imbalance. Additionally, this metric can be extended to include unlabeled and synthetic nodes, which has been proven beneficial for augmenting the training set and addressing class imbalance issues~\citep{zhao2021graphsmote,wu2021graphmixup,Zhou_Gong_2023graphsr}. To utilize this metric, we need to generate pseudo-labels for unlabeled nodes. As for synthetic nodes, we synthesize their input features, edge connections, and labels. A key advantage of our metric lies in its independence from specific methods for generating pseudo-labels and synthesizing nodes, ensuring compatibility with any existing techniques in the literature. In this work, we demonstrate that even the simplest methods yield superior performance when equipped with our metric. In the subsequent sections, we describe how we incorporate valuable labeled, unlabeled, and synthetic nodes into the training set, each of which forms a distinct component. 

\paragraph{Component 1} The first component we construct is to identify labeled nodes that are worth training in a class-imbalanced setting using our importance metric. Let $\gD_{l}=\{(v, y_v)\}$ denote the set of labeled nodes. The set of valuable labeled nodes $\tilde{D}_{l}$ is defined as:
\begin{equation} \label{eq:valuable_labeled}
\begin{aligned}
    \tilde{D}_{l} = \{(v, y_v) \in D_{l}\ |\ \eta_v >0 \}.
\end{aligned}
\end{equation}

\paragraph{Component 2} As the labels for the unlabeled nodes are unavailable during training, we generate pseudo-labels for them before applying the importance metric. In this study, we choose the class with the highest prediction probability from the model as the pseudo-label. For each unlabeled node $v \in \gD_{ul}$, its pseudo-label $\hat{y}_v$ is:
\begin{equation} \label{eq:pesudo-label}
\begin{aligned}
    \hat{y}_v = \arg\max_{k} h_{v}[k],
\end{aligned}
\end{equation}
where $h_{v}=H[v,:] \in\mathbb{R}^{1\times c}$ is the prediction probability distribution of the current GNN model on node $v$. With the pseudo-label $\hat{y}_v$, we replace $y_v$ in Eq.\ref{eq:final_expression} with it to compute the importance $\eta_v$ for the unlabeled node. The filtered set of unlabeled nodes $\tilde{D}_{ul}$ is then:
\begin{equation} \label{eq:valuable_unlabeled}
\begin{aligned}
    \tilde{D}_{ul} = \{(v, \hat{y}_{v})\ |\ v \in D_{ul} \text{ and } \eta_v >0 \}.
\end{aligned}
\end{equation}

\paragraph{Component 3} In this study, we adopt a MixUp-like methodology to generate synthetic nodes, augmenting the training dataset to mitigate class imbalance. Firstly, we draw a set of node pairs $\gS=\{\langle(v_s,y_s), (v_t, y_t)\rangle\}$ from the training set. Specifically, nodes \(v_s\) are sampled until each class reaches the maximum sample count among all classes in the training set, while nodes \(v_t\) are sampled based on the probability \(p_t(u)\) defined as follows:
\begin{equation} \label{eq:sample_probability}
\begin{aligned}
    p_t(u) = \frac{\text{log }(|\gV_k|+1)}{(|\gV_k|+1)\sum_{j=1}^{c}\text{log }(|\gV_j|+1)}.
\end{aligned}
\end{equation}
where \(k\) is the label of node \(u\) and \(|\gV_k|\) stands for the subset of nodes with label \(k\). This increases the likelihood of selecting nodes from minority classes for synthesis, thereby enriching their patterns within the training set. For each node pair $\langle (v_s, y_s), (v_t, y_t) \rangle \in \gS$, we generate a new node $v_{syn}$ by employing MixUp \cite{Zhang2019mixup} from three aspects, namely feature, neighborhood, and label mixing:
\begin{equation} \label{eq.synthetic_formula}
    \begin{cases}
      x_{syn} = \lambda \cdot x_s + (1-\lambda) \cdot x_t,\\
      {A'}_{syn} = \lambda \cdot A_{s} + (1-\lambda) \cdot A_{t}, \\
      y_{syn} = \lambda \cdot y_s + (1-\lambda) \cdot y_t.
    \end{cases}       
\end{equation}
where $\lambda \sim \text{Beta}(\alpha,\alpha)$ represents the synthetic scale drawn from the Beta distribution with the parameter $\alpha$, and $A'_{syn}\in\mathbb{R}^{1\times n}$ denotes the combined adjacency matrix for nodes $v_s$ and $v_t$. Since $A'_{syn}$ consists of real numbers, it must be converted into a discrete version. This involves sampling a specific number of neighbors for node $v_{syn}$ based on $A'_{syn}$, resulting in a discrete form $A_{syn}$:
\begin{equation}
\begin{aligned}
    A_{syn} = \text{Sample} (A'_{syn}),
\end{aligned}
\end{equation}
where the probability of an item $A_{syn}[k]$ being one is proportional to the corresponding value in $A'_{syn}[k]$. To maintain consistent degree statistics as per common practice in prior research~\citep{park2022graphens,li2023graphsha}, the number of sampled neighbors is drawn from a separate degree distribution of the entire graph. In this way, we produce a set of new nodes, \ie, $\gD_{syn}=\{(v_{syn},y_{syn})\}$, based on the sampled node pairs $\gS$. Combining Eq.~\ref{eq:final_expression} and Eq.~\ref{eq.synthetic_formula}, we calculate the importance of a synthetic node $\eta_{v_{syn}}$ to determine whether to use it for training. Finally, we build a set of valuable synthetic nodes:
\begin{equation} \label{eq:valuable_synthetic}
\begin{aligned}
    \tilde{D}_{syn} = \{(v_{syn}, {y}_{syn})\ \in \gD_{syn}\ |\ \eta_{v_{syn}} >0 \}.
\end{aligned}
\end{equation}

\stitle{Training objective.} The overall training objective $\gL$ is defined as the sum of the cross entropy loss on high-quality labeled nodes $\gL_{l}$, unlabeled nodes $\gL_{ul}$, and synthetic nodes $\gL_{syn}$:
\begin{equation} \label{eq:final_loss}
\begin{aligned}
    \gL = \gL_{l} + \beta \cdot \gL_{ul} + \gamma \cdot \gL_{syn},
\end{aligned}
\end{equation}
where $\beta$ and $\gamma$ are scaling factors for the losses of the unlabeled and synthetic nodes, respectively.

\subsection{Meta-set construction}
\label{sec:meta-set-construction}
The effectiveness of our framework is heavily dependent on the quality of the meta-set due to its small size. Our goal is to select a fixed number of nodes for each class \( k \) that can accurately reflect the distribution of the initial training pool \( \mathcal{V}_k \). Let \( \mathcal{M}_k \subseteq \mathcal{V}_k \) denote the set of meta-samples for class \( k \). We aim to find the optimal set \( \tilde{\mathcal{M}}_k \) by minimizing the following objective:
\begin{equation}
\begin{aligned}
 \textstyle \tilde{\gM}_k = \arg\min_{\gM_k} \frac{1}{|\gV_k|} \sum_{v\in\gV_k} D(v, \gM_k),
\end{aligned}
\end{equation}
where $D(\cdot,\cdot)$ represents the distance between a node $v$ and the meta-set $\gM_k$. To calculate this distance, we exploit the context embedding for the nodes, $F=\tilde{A} X$, which captures not only the nodes' features but also their local structures. Let $f_v$ be the context embedding vector of a node $v$, and we define the distance $D(v, \gM_k)$ between the node and the meta-set $\gM_k$ as:
\begin{equation}
\textstyle     \begin{aligned}
   D(v,\gM_k) = \min_{u \in \gM_k} \text{dist}(f_v, f_u),
    \end{aligned}
\end{equation}
where $\text{dist}(\cdot, \cdot)$ denotes the distance metric between two vectors. In this work, this metric can be the Euclidean distance or the Manhattan distance. Intuitively, the distance between a node and the meta-set is formulated as the distance to its closest meta-sample in the meta-set. Given the impracticality of finding the optimal solution through exhaustive search, we employ a heuristic search method. In particular, we utilize the Partitioning Around Medoids (PAM) algorithm~\citep{park2009simple} to identify meta-samples for each class. The meta-set utilized is then the union of meta-samples from each class:
\begin{equation}
\begin{aligned}
\textstyle  \gD_{meta} = \{ (v,y_v)\ |\ v \in \bigcup_{k=1}^{c} \tilde{\gM}_k \}.
\end{aligned}
\end{equation}
In the end, we exclude the meta-set $\gD_{meta}$ from the training set $\gD_{tr}$ to form the set of labeled nodes $\gD_{l}$.

\section{Experiments}




\subsection{Experimental Setup}

\begin{table}[t]
  \caption{Statistics of datasets adopted in this work}
  \label{table:data_statistics}
  \centering
  \begin{tabular}{crrrr}
  \toprule
{Dataset}  & {Nodes} & {Edges} & {Features} & {Classes} \\
\midrule
Cora & 2,708 & 10,556 & 1,433 & 7 \\
CiteSeer & 3,327 & 9,104 & 3,703 & 6 \\
PubMed & 19,717 & 88,648 & 500 & 3 \\
Amazon-Photo  & 7,650 & 119,081 & 745 & 8 \\
Amazon-Computers & 13,752 & 245,861 & 767 & 10 \\
\bottomrule
\end{tabular}
\end{table}

\begin{table*}[t]
\caption{Performance comparison (\%) on benchmark graphs with an imbalance ratio of 50. The best results are in bold. Models with performances within the standard error are statistically comparable.}
\begin{center}
\setlength{\columnsep}{1pt}%
\resizebox{0.99\linewidth}{!}{
\begin{tabular}{@{\extracolsep{0pt}}rlcccccccccc@{}}
\toprule
 & \multirow{1}{*}{\textbf{Dataset}} & \multicolumn{2}{c}{Cora} & \multicolumn{2}{c}{CiteSeer} & \multicolumn{2}{c}{PubMed} & \multicolumn{2}{c}{Photo} & \multicolumn{2}{c}{Computers} \\ 
 
& \textbf{(IR=50)} & bAcc. & Macro F1 & bAcc. & Macro F1 & bAcc. & Macro F1 & bAcc. & Macro F1 & bAcc. & Macro F1 \\
\midrule

\multirow{14}{*}{\rotatebox[origin=c]{90}{GCN}} 

& Vanilla & \ms{{75.38 }}{0.67} & \ms{{77.98 }}{0.65} & \ms{51.78}{0.88} & \ms{49.14}{1.37} & \ms{52.08}{0.67} & \ms{48.31}{1.02} & \ms{78.08}{0.17} & \ms{78.44}{0.12} & \ms{66.91}{1.29} & \ms{68.18}{1.39} \\

& GRAND & \ms{71.55}{0.82} & \ms{74.35}{0.92} & \ms{47.41}{0.78} & \ms{43.78}{1.20} & \ms{46.30}{0.80} & \ms{39.86}{0.91} & \ms{78.67}{0.15} & \ms{79.05}{0.14} & \ms{70.78}{0.40} & \ms{72.26}{0.40} \\

\cmidrule{2-12}

& ReWeight & \ms{81.65}{0.72} & \ms{81.85}{0.66} & \ms{60.99}{0.86} & \ms{60.19}{0.97} & \ms{81.06}{0.37} & \ms{80.90}{0.31} & \ms{90.45}{0.04} & \ms{90.28}{0.10} & \ms{87.76}{0.17} & \ms{84.74}{0.16} \\

& PC Softmax & \ms{81.68}{0.70} & \ms{79.68}{0.62} & \ms{62.10}{0.84} & \ms{62.24}{0.81} & \ms{79.81}{0.43} & \ms{76.69}{0.63} & \ms{89.76}{0.28} & \ms{88.13}{0.24} & \ms{85.84}{0.78} & \ms{82.29}{0.67} \\

& Balanced Softmax & \ms{81.90}{0.64} & \ms{81.51}{0.61} & \ms{{\bf 64.93}}{0.67} & \ms{{\bf 64.86}}{0.66} & \ms{82.37}{0.33} & \ms{81.65}{0.26} & \ms{90.02}{0.11} & \ms{88.67}{0.19} & \ms{86.27}{0.35} & \ms{81.76}{0.64} \\

& TAM(BS) & \ms{82.42}{0.63} & \ms{82.12}{0.55} & \ms{{\bf 64.94}}{0.64} & \ms{{\bf 64.89}}{0.65} & \ms{81.54}{0.38} & \ms{80.64}{0.37} & \ms{86.19}{0.12} & \ms{82.87}{0.10} & \ms{77.76}{1.42} & \ms{71.50}{1.20} \\

& ReNode & \ms{81.19}{0.61} & \ms{81.82}{0.53} & \ms{60.50}{0.79} & \ms{59.64}{0.92} & \ms{80.90}{0.38} & \ms{80.61}{0.38} & \ms{90.27}{0.08} & \ms{90.01}{0.04} & \ms{87.84}{0.08} & \ms{85.06}{0.15} \\

& TAM(ReNode) & \ms{80.56}{0.72} & \ms{81.79}{0.65} & \ms{60.74}{0.64} & \ms{59.92}{0.83} & \ms{80.65}{0.66} & \ms{81.05}{0.64} & \ms{88.31}{0.08} & \ms{88.08}{0.10} & \ms{84.34}{0.25} & \ms{82.21}{0.19} \\

& \Nan{ReVar} & \Nan{\ms{75.01}{0.66}} & \Nan{\ms{77.52}{0.70}} & \Nan{\ms{61.10}{0.95}} & \Nan{\ms{59.01}{1.27}} & \Nan{\ms{70.93}{0.93}} & \Nan{\ms{71.25}{0.83}} & \Nan{\ms{81.17}{0.11}} & \Nan{\ms{80.36}{0.09}} & \Nan{\ms{85.11}{0.40}} & \Nan{\ms{86.02}{0.24}} \\

& \Nan{BAT} & \Nan{\ms{76.70}{0.47}} & \Nan{\ms{79.21}{0.48}} & \Nan{\ms{54.39}{1.17}} & \Nan{\ms{52.38}{1.65}} & \Nan{\ms{52.50}{0.83}} & \Nan{\ms{48.70}{1.22}} & \Nan{\ms{78.40}{0.21}} & \Nan{\ms{78.83}{0.18}} & \Nan{\ms{66.01}{1.32}} & \Nan{\ms{67.02}{1.50}} \\

\cmidrule{2-12}

& GraphENS & \ms{82.17}{0.55} & \ms{81.73}{0.49} & \ms{63.07}{0.82} & \ms{62.38}{0.96} & \ms{81.62}{0.36} & \ms{81.22}{0.30} & \ms{91.25}{0.07} & \ms{90.54}{0.11} & \ms{86.82}{0.15} & \ms{84.07}{0.10} \\

& TAM(G-ENS) & \ms{82.08}{0.56} & \ms{81.96}{0.44} & \ms{63.59}{0.78} & \ms{63.09}{0.95} & \ms{82.63}{0.30} & \ms{82.10}{0.29} & \ms{88.55}{0.12} & \ms{86.67}{0.20} & \ms{84.18}{0.20} & \ms{81.84}{0.22} \\

& GraphSHA & \ms{{\bf 83.42}}{0.54} & \ms{82.80}{0.50} & \ms{62.25}{0.87} & \ms{62.32}{0.97} & \ms{79.17}{0.37} & \ms{79.46}{0.35} & \ms{88.34}{0.35} & \ms{88.55}{0.22} & \ms{83.31}{0.16} & \ms{83.09}{0.13} \\

\cmidrule{2-12}

& \model & \ms{{ \bf 83.71 }}{0.49} & \ms{{ \bf 83.24 }}{0.51} & \ms{{ \bf 65.13 }}{0.69} & \ms{{\bf 64.28}}{0.78} & \ms{{ \bf 84.00 }}{0.25} & \ms{{ \bf 83.65 }}{0.21} & \ms{{ \bf 91.75 }}{0.09} & \ms{{ \bf 92.00 }}{0.05} & \ms{{ \bf 90.77 }}{0.04} & \ms{{ \bf 87.91 }}{0.04} \\

\cline{1-12}
\noalign{\vskip\doublerulesep
         \vskip-\arrayrulewidth} \cline{1-12}
\rule{0pt}{2.5ex}  

\multirow{14}{*}{\rotatebox[origin=c]{90}{GAT}} 

& Vanilla & \ms{61.58}{1.59} & \ms{63.90}{1.77} & \ms{48.65}{0.78} & \ms{45.37}{1.27} & \ms{45.56}{1.47} & \ms{38.52}{1.88} & \ms{70.26}{0.99} & \ms{70.22}{1.13} & \ms{63.14}{1.02} & \ms{64.30}{1.32} \\

& GRAND & \ms{56.28}{1.29} & \ms{58.58}{1.29} & \ms{42.91}{0.87} & \ms{38.60}{0.97} & \ms{44.22}{1.35} & \ms{36.98}{1.78} & \ms{72.43}{0.80} & \ms{72.97}{0.81} & \ms{64.55}{1.19} & \ms{66.32}{1.29} \\

\cmidrule{2-12}

& ReWeight & \ms{{\bf 82.09}}{0.69} & \ms{81.34}{0.61} & \ms{60.78}{0.92} & \ms{60.17}{0.97} & \ms{79.58}{0.48} & \ms{79.65}{0.29} & \ms{84.13}{0.40} & \ms{83.93}{0.33} & \ms{78.99}{0.33} & \ms{78.44}{0.39} \\

& PC Softmax & \ms{74.91}{0.99} & \ms{74.56}{1.00} & \ms{62.43}{0.89} & \ms{62.25}{0.91} & \ms{79.81}{0.29} & \ms{79.12}{0.21} & \ms{83.23}{0.57} & \ms{81.66}{0.53} & \ms{76.44}{0.66} & \ms{72.47}{0.79} \\

& Balanced Softmax & \ms{75.19}{1.40} & \ms{74.49}{1.24} & \ms{62.05}{0.47} & \ms{62.16}{0.49} & \ms{80.68}{0.35} & \ms{80.13}{0.44} & \ms{85.33}{0.53} & \ms{84.31}{0.48} & \ms{79.25}{0.19} & \ms{76.33}{0.38} \\

& TAM(BS) & \ms{74.43}{0.90} & \ms{74.36}{0.99} & \ms{63.50}{0.63} & \ms{{\bf 63.53}}{0.66} & \ms{80.86}{0.48} & \ms{80.56}{0.41} & \ms{81.69}{0.57} & \ms{79.52}{0.51} & \ms{76.40}{1.13} & \ms{72.06}{1.50} \\

& ReNode & \ms{79.82}{1.00} & \ms{80.38}{0.86} & \ms{57.94}{0.72} & \ms{56.80}{0.79} & \ms{79.15}{0.51} & \ms{79.26}{0.33} & \ms{83.31}{0.40} & \ms{83.71}{0.48} & \ms{78.65}{0.32} & \ms{78.14}{0.31} \\

& TAM(ReNode) & \ms{77.96}{0.87} & \ms{79.07}{0.75} & \ms{57.71}{0.87} & \ms{56.30}{0.96} & \ms{79.44}{1.15} & \ms{79.68}{1.02} & \ms{80.54}{0.60} & \ms{80.33}{0.65} & \ms{77.92}{0.44} & \ms{76.61}{0.36} \\

& \Nan{ReVar} & \Nan{\ms{75.82}{0.63}} & \Nan{\ms{78.08}{0.64}} & \Nan{\ms{61.67}{0.94}} & \Nan{\ms{59.61}{1.30}} & \Nan{\ms{66.77}{1.79}} & \Nan{\ms{66.56}{2.47}} & \Nan{\ms{81.67}{0.02}} & \Nan{\ms{80.52}{0.02}} & \Nan{\ms{83.28}{0.76}} & \Nan{\ms{84.71}{0.72}} \\

& \Nan{BAT} & \Nan{\ms{66.20}{1.67}} & \Nan{\ms{68.48}{1.83}} & \Nan{\ms{50.09}{0.98}} & \Nan{\ms{47.59}{1.47}} & \Nan{\ms{45.48}{1.50}} & \Nan{\ms{38.40}{1.93}} & \Nan{\ms{72.31}{1.00}} & \Nan{\ms{72.50}{1.03}} & \Nan{\ms{63.54}{1.52}} & \Nan{\ms{64.52}{1.93}} \\

\cmidrule{2-12}

& GraphENS & \ms{80.20}{0.80} & \ms{80.14}{0.79} & \ms{61.72}{0.61} & \ms{61.14}{0.69} & \ms{81.06}{0.32} & \ms{80.73}{0.29} & \ms{88.81}{0.19} & \ms{88.94}{0.16} & \ms{87.39}{0.32} & \ms{85.33}{0.16} \\

& TAM(G-ENS) & \ms{81.25}{0.62} & \ms{81.45}{0.59} & \ms{62.11}{0.74} & \ms{61.42}{0.91} & \ms{82.33}{0.27} & \ms{81.75}{0.25} & \ms{86.99}{0.16} & \ms{86.13}{0.22} & \ms{85.19}{0.23} & \ms{82.76}{0.16} \\

& GraphSHA & \ms{81.22}{0.66} & \ms{81.07}{0.64} & \ms{60.51}{0.84} & \ms{60.53}{0.95} & \ms{80.59}{0.33} & \ms{80.77}{0.25} & \ms{85.93}{0.30} & \ms{87.07}{0.27} & \ms{83.29}{0.32} & \ms{83.11}{0.28} \\

\cmidrule{2-12}

& \model & \ms{{ \bf 82.60 }}{0.48} & \ms{{ \bf 82.55 }}{0.43} & \ms{{ \bf 64.78 }}{0.62} & \ms{{ \bf 63.38 }}{0.73} & \ms{{ \bf 84.07 }}{0.18} & \ms{{ \bf 83.28 }}{0.25} & \ms{{ \bf 92.00 }}{0.08} & \ms{{ \bf 91.81 }}{0.17} & \ms{{ \bf 90.50 }}{0.09} & \ms{{ \bf 88.76 }}{0.04} \\

\cline{1-12}
\noalign{\vskip\doublerulesep
         \vskip-\arrayrulewidth} \cline{1-12}
\rule{0pt}{2.5ex}  

\multirow{14}{*}{\rotatebox[origin=c]{90}{GraphSAGE}} 

& Vanilla & \ms{71.99}{0.72} & \ms{75.19}{0.72} & \ms{{48.20 }}{1.07} & \ms{{45.70 }}{1.46} & \ms{59.60}{0.63} & \ms{58.04}{0.96} & \ms{{84.87 }}{0.49} & \ms{{86.95 }}{0.46}  & \ms{77.28}{0.93} & \ms{78.97}{0.94} \\

& GRAND & \ms{62.39}{1.00} & \ms{65.12}{1.17} & \ms{43.41}{0.87} & \ms{39.99}{1.19} & \ms{54.53}{0.26} & \ms{48.86}{0.30} & \ms{77.32}{0.85} & \ms{77.85}{0.83} & \ms{71.78}{0.60} & \ms{73.11}{0.63} \\

\cmidrule{2-12}

& ReWeight & \ms{78.34}{0.81} & \ms{79.85}{0.65} & \ms{55.56}{0.93} & \ms{54.74}{0.95} & \ms{76.88}{0.37} & \ms{77.41}{0.35} & \ms{90.26}{0.14} & \ms{91.39}{0.07} & \ms{86.91}{0.13} & \ms{86.39}{0.08} \\

& PC Softmax & \ms{81.16}{0.49} & \ms{80.37}{0.40} & \ms{60.69}{0.92} & \ms{60.56}{0.95} & \ms{81.42}{0.28} & \ms{80.37}{0.24} & \ms{91.97}{0.12} & \ms{91.89}{0.09} & \ms{87.67}{0.15} & \ms{85.09}{0.21} \\

& Balanced Softmax & \ms{80.71}{0.70} & \ms{80.73}{0.59} & \ms{59.82}{1.19} & \ms{59.94}{1.25} & \ms{79.70}{0.42} & \ms{78.87}{0.40} & \ms{92.10}{0.21} & \ms{91.57}{0.11} & \ms{87.38}{0.33} & \ms{84.37}{0.21} \\

& TAM(BS) & \ms{80.73}{0.51} & \ms{81.06}{0.43} & \ms{59.44}{1.07} & \ms{59.61}{1.06} & \ms{78.94}{0.44} & \ms{78.50}{0.43} & \ms{87.22}{0.37} & \ms{85.08}{0.24} & \ms{85.15}{0.48} & \ms{81.12}{0.39} \\

& ReNode & \ms{78.07}{0.60} & \ms{79.89}{0.53} & \ms{55.07}{1.20} & \ms{54.22}{1.18} & \ms{75.72}{0.16} & \ms{76.52}{0.17} & \ms{89.79}{0.13} & \ms{90.95}{0.08} & \ms{87.13}{0.12} & \ms{86.45}{0.03} \\

& TAM(ReNode) & \ms{78.93}{0.67} & \ms{80.80}{0.65} & \ms{55.52}{0.98} & \ms{54.78}{0.98} & \ms{76.30}{0.57} & \ms{76.95}{0.56} & \ms{88.00}{0.18} & \ms{89.10}{0.14} & \ms{84.75}{0.27} & \ms{84.36}{0.17} \\

& \Nan{ReVar} & \Nan{\ms{66.80}{0.64}} & \Nan{\ms{66.80}{0.71}} & \Nan{\ms{57.70}{0.87}} & \Nan{\ms{55.01}{1.23}} & \Nan{\ms{75.98}{0.97}} & \Nan{\ms{75.89}{0.71}} & \Nan{\ms{85.72}{0.03}} & \Nan{\ms{87.56}{0.03}} & \Nan{\ms{82.45}{0.36}} & \Nan{\ms{84.34}{0.19}} \\

& \Nan{BAT} & \Nan{\ms{77.58}{0.64}} & \Nan{\ms{79.77}{0.55}} & \Nan{\ms{51.17}{1.09}} & \Nan{\ms{49.69}{1.51}} & \Nan{\ms{59.69}{0.60}} & \Nan{\ms{58.51}{0.84}} & \Nan{\ms{83.15}{0.18}} & \Nan{\ms{84.59}{0.26}} & \Nan{\ms{78.56}{0.45}} & \Nan{\ms{80.35}{0.49}} \\

\cmidrule{2-12}

& GraphENS & \ms{80.40}{0.66} & \ms{80.44}{0.60} & \ms{62.20}{0.85} & \ms{61.33}{0.98} & \ms{82.30}{0.19} & \ms{82.20}{0.11} & \ms{91.70}{0.13} & \ms{92.06}{0.07} & \ms{87.94}{0.20} & \ms{85.92}{0.07} \\

& TAM(G-ENS) & \ms{81.19}{0.49} & \ms{81.35}{0.45} & \ms{{63.44}}{0.75} & \ms{{\bf 62.65}}{0.92} & \ms{82.29}{0.18} & \ms{82.04}{0.17} & \ms{90.23}{0.29} & \ms{89.62}{0.16} & \ms{86.75}{0.22} & \ms{84.40}{0.15} \\

& GraphSHA & \ms{80.29}{0.29} & \ms{80.09}{0.47} & \ms{57.69}{1.18} & \ms{57.78}{1.32} & \ms{75.40}{0.46} & \ms{75.77}{0.28} & \ms{90.56}{0.17} & \ms{91.68}{0.10} & \ms{87.75}{0.17} & \ms{{\bf 86.93}}{0.07} \\

\cmidrule{2-12}

& \model & \ms{{ \bf 82.90 }}{0.41} & \ms{{ \bf 83.28 }}{0.37} & \ms{{\bf 64.22 }}{0.62} & \ms{{\bf 62.89 }}{0.78} & \ms{{ \bf 86.25 }}{0.24} & \ms{{ \bf 85.90 }}{0.17} & \ms{{\bf 93.02 }}{0.15} & \ms{{ \bf 92.71 }}{0.10} & \ms{{ \bf 88.94 }}{0.22} & \ms{{ \bf 86.92 }}{0.11} \\ 

\bottomrule
\end{tabular}
}
\end{center}
\label{table:main_imb_50}
\end{table*}

\stitle{Datasets.} 
We adopt five benchmark datasets, which consist of three citation networks (Cora, CiteSeer, and PubMed)~\citep{yang2016revisiting} and two Amazon co-purchase networks (Amazon-Photo and Amazon-Computers)~\citep{shchur2018pitfalls}. Dataset statistics are summarized in Table~\ref{table:data_statistics}. For all datasets, we set the imbalance ratio $IR$ to 50 and construct a long-tailed training set following ~\citep{cui2019class}. For the citation datasets, we use the public data splits from ~\citep{pei2020geom} and iteratively remove nodes from the training set to create a long-tailed distribution. For the co-purchase networks, we set the training ratio to 10\% and sample nodes for each class to follow a long-tailed distribution. The remaining data is split into a 1:8 ratio for validation and testing. Unlike the citation datasets, the co-purchase datasets are naturally imbalanced, meaning that their validation and test sets are also highly imbalanced. More details on the data sets are available in Appendix~\ref{desc-datasets}.

\stitle{Baselines.} We compare our model against eleven established baselines, which are categorized into three groups: (1) \emph{Traditional methods}, which include the vanilla approach using conventional cross-entropy loss and GRAND~\cite{feng2020grand}, which utilizes unlabeled nodes to enhance model training; (2) \emph{Algorithm-level methods}, consisting of Re-weight~\cite{japkowicz2002class}, Balanced Softmax~\cite{ren2020balanced}, PC Softmax~\cite{hong2021disentangling}, ReNode~\cite{chen2021topology}, TAM~\cite{song2022tam}, \Nan{ReVar~\citep{yan2023rethinking}, and BAT~\citep{Liu24BAT}}; (3) \emph{Data-level methods}, including GraphENS~\cite{park2022graphens} and GraphSHA~\cite{li2023graphsha}. To ensure a comprehensive evaluation, we integrate TAM with Balanced Softmax, ReNode, and GraphENS. Detailed descriptions of baselines and their hyper-parameter configurations are in Appendix~\ref{desc-baselines}.

\stitle{Evaluation protocol.} In this study, we deploy both our proposed model and the selected baselines across three representative GNN architectures: GCN~\citep{kipf2017semisupervised}, GAT~\cite{velivckovic2017graph}, and GraphSAGE~\cite{hamilton2017inductive}. Nodes that are not part of the training set are treated as unlabeled in our setting. For the evaluation of model performance, we employ three metrics: Accuracy (Acc.), balanced Accuracy (BAcc.), and the Macro F1 score (Macro F1). All these metrics are scaled between 0 and 1, where higher values represent better model performance. We conduct ten independent runs for each method and report the average score and the standard error for a robust statistical analysis. More details about the evaluation process are provided in Appendix~\ref{desc-evaluation}.

\subsection{Model Performance} \label{sec:model_performance}

\stitle{Overall performance.} The results in Table~\ref{table:main_imb_50} demonstrate that our model consistently surpasses the current state-of-the-art methods across various GNN backbones, achieving significant improvements. This can be attributed to our model's capacity to identify valuable unlabeled nodes for enhancing the training set and effectively address class imbalance by integrating high-quality synthetic nodes into minority classes. Notably, our model's performance remains consistent across different GNN architectures, indicating reduced dependency on the underlying GNN structures.

Furthermore, our evaluation reveals that node oversampling strategies, such as GraphENS~\citep{park2022graphens} and GraphSHA~\citep{li2023graphsha}, generally yield better results compared to algorithm-level techniques. Among algorithm-level approaches, Balanced Softmax~\citep{ren2020balanced} and PC Softmax~\cite{hong2021disentangling} prove to be effective in most scenarios. When comparing GRAND~\cite{feng2020grand} to the vanilla approach, we observe that leveraging unlabeled nodes can boost performance in some cases, highlighting their potential to mitigate class imbalance. However, in most instances, the vanilla approach outperforms GRAND, suggesting that careful handling of unlabeled nodes is crucial under class imbalance conditions to avoid negative impacts on model performance. The results in terms of accuracy are deferred to Appendix~\ref{appendix-overall-acc}.

\begin{figure}[!ht]
  \centering
  \includegraphics[width=\linewidth]{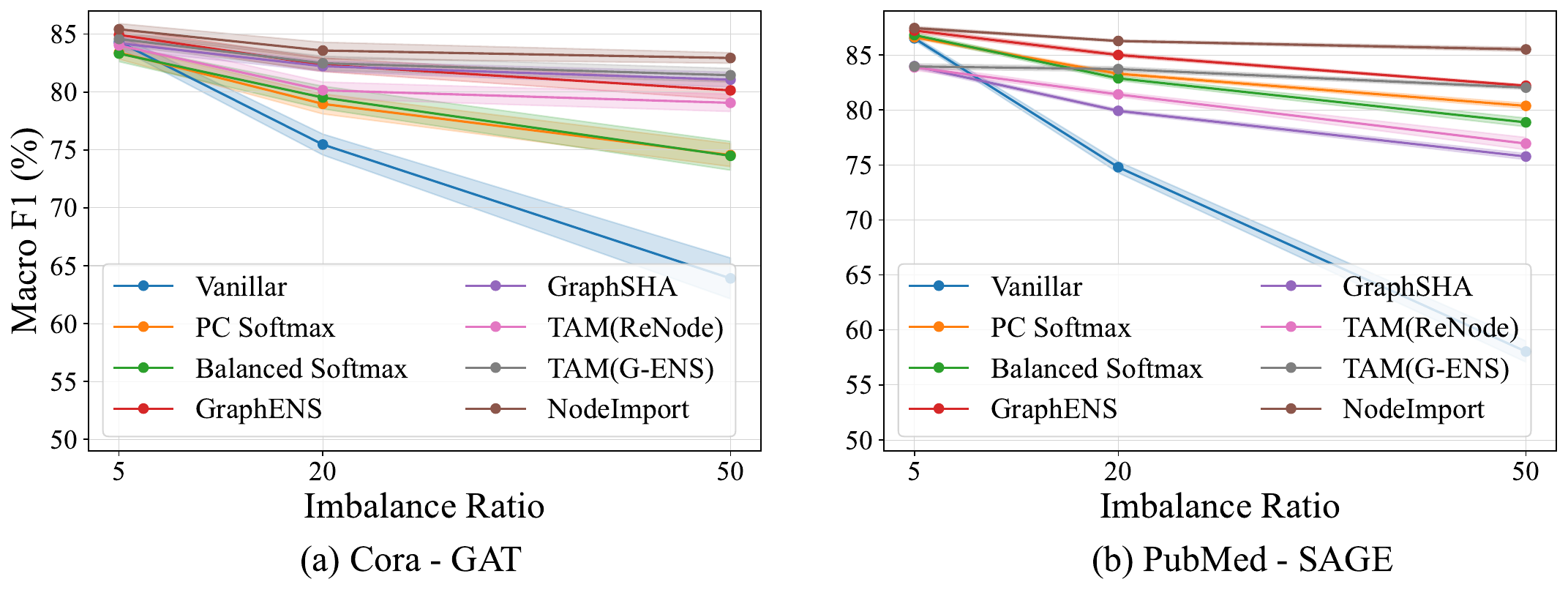}
  \caption{\Nan{Model performance under varying imbalance ratios in terms of Macro F1 score.}
  \Description{Model performance under varying ratios.}}
  \label{figure:main-imb-trend}
\end{figure}

\stitle{Performance under different imbalance ratios.} Figure~\ref{figure:main-imb-trend} illustrates the behavior of various methods under different imbalance ratios ($\text{IR}\in\{5,20,50\}$). As the imbalance ratio decreases, the performance gap in Macro F1 scores among different methods narrows. Conversely, as the imbalance ratio increases, our model's superiority becomes more pronounced, demonstrating its effectiveness in handling highly imbalanced problems. Even with a small imbalance ratio, our model delivers slight performance improvements, likely due to the integration of valuable unlabeled and synthetic nodes, which helps regularize model training. Results for other metrics, which are available in Appendix~\ref{appendix-performance-trend}, exhibit a similar trend.

\subsection{Model Analysis}

We conduct a comprehensive analysis of various facets of \model. However, due to space limitations, here we focus on presenting the ablation study, analysis of the meta-set construction method, and a case study on per-class F1 scores. 
For further details, additional experimental results are provided in Appendix~\ref{app.add-experiments}, including evaluations on a large-scale dataset (Appendix~\ref{app.large-scale-dataset}), a sensitivity analysis of hyper-parameters (Appendix~\ref{app.parameters-sensitivity}), and a case study on the class distribution of filtered labeled nodes (Appendix~\ref{app.class-distribution}).


\begin{table}[!ht]
\centering
\caption{Ablation study of model components. {\cmark} indicates the inclusion of a component, whereas {\grayx} denotes its exclusion. }
\setlength{\tabcolsep}{2pt}
\resizebox{0.98\linewidth}{!}{
\begin{tabular}{cccc@{\hskip 5pt}c@{\hskip 10pt}c@{\hskip 10pt}c}
\toprule
\textbf{} & \textbf{C1} & \textbf{C2} & \textbf{C3} & Acc. & bAcc. & Macro F1  \\

\midrule

\multirow{8}{*}{\shortstack[1]{Cora \\ + \\ GCN}}& \grayx& \grayx& \grayx & \ms{{81.55 }}{0.51} & \ms{{75.38 }}{0.67} & \ms{{77.98 }}{0.65} \\ 

& \cmark& \grayx& \grayx & \ms{{83.70 }}{0.61} & \ms{{79.92 }}{0.80} & \ms{{81.12 }}{0.70} \\ 

& \grayx& \cmark& \grayx & \ms{{84.12 }}{0.28} & \ms{{81.13 }}{0.13} & \ms{{82.01 }}{0.27} \\ 

& \grayx& \grayx& \cmark & \ms{{83.72 }}{0.56} & \ms{{80.39 }}{0.73} & \ms{{81.58 }}{0.60} \\ 

& \grayx& \cmark& \cmark & \ms{{\bf 85.33 }}{0.33} & \ms{{\bf 83.85 }}{0.53} & \ms{{\bf 83.50 }}{0.46} \\ 

& \cmark& \grayx& \cmark & \ms{{83.82 }}{0.60} & \ms{{81.36 }}{0.71} & \ms{{82.05 }}{0.60} \\ 

& \cmark& \cmark& \grayx & \ms{{84.53 }}{0.41} & \ms{{82.56 }}{0.43} & \ms{{82.48 }}{0.44} \\ 

& \cmark& \cmark& \cmark & \ms{{ \bf 85.11 }}{0.42} & \ms{{ \bf 83.71 }}{0.49} & \ms{{ \bf 83.24 }}{0.51} \\

\midrule

\multirow{8}{*}{\shortstack[1]{CiteSeer \\ + \\ GraphSAGE}}& \grayx& \grayx& \grayx & \ms{{56.07 }}{1.65} & \ms{{48.20 }}{1.07} & \ms{{45.70 }}{1.46} \\ 

& \cmark& \grayx& \grayx & \ms{{62.71 }}{1.18} & \ms{{55.67 }}{0.80} & \ms{{53.95 }}{1.02} \\ 

& \grayx& \cmark& \grayx & \ms{{66.65 }}{1.12} & \ms{{59.99 }}{0.99} & \ms{{57.91 }}{1.05} \\ 

& \grayx& \grayx& \cmark & \ms{{64.01 }}{1.35} & \ms{{56.44 }}{0.83} & \ms{{54.68 }}{1.06} \\ 

& \grayx& \cmark& \cmark & \ms{{\bf 70.45 }}{1.04} & \ms{{63.42 }}{0.95} & \ms{{61.91 }}{1.17} \\ 

& \cmark& \grayx& \cmark & \ms{{65.30 }}{1.13} & \ms{{58.14 }}{0.86} & \ms{{56.56 }}{1.04} \\ 

& \cmark& \cmark& \grayx & \ms{{68.99 }}{0.87} & \ms{{62.31 }}{0.54} & \ms{{60.98 }}{0.69} \\ 

& \cmark& \cmark& \cmark & \ms{{\bf 71.12 }}{0.84} & \ms{{\bf 64.22 }}{0.62} & \ms{{\bf 62.89 }}{0.78} \\

\midrule

\multirow{8}{*}{\shortstack[1]{photo \\ + \\ GraphSAGE}}& \grayx& \grayx& \grayx & \ms{{90.53 }}{0.30} & \ms{{84.87 }}{0.49} & \ms{{86.95 }}{0.46} \\ 

& \cmark& \grayx& \grayx & \ms{{90.49 }}{0.12} & \ms{{84.59 }}{0.41} & \ms{{86.04 }}{0.43} \\ 

& \grayx& \cmark& \grayx & \ms{{91.16 }}{0.30} & \ms{{88.42 }}{0.39} & \ms{{89.58 }}{0.32} \\ 

& \grayx& \grayx& \cmark & \ms{{90.92 }}{0.11} & \ms{{88.01 }}{0.63} & \ms{{88.43 }}{0.45} \\ 

& \grayx& \cmark& \cmark & \ms{{93.78 }}{0.04} & \ms{{92.56 }}{0.18} & \ms{{92.49 }}{0.08} \\ 

& \cmark& \grayx& \cmark & \ms{{91.11 }}{0.09} & \ms{{88.64 }}{0.43} & \ms{{88.99 }}{0.18} \\ 

& \cmark& \cmark& \grayx & \ms{{91.36 }}{0.42} & \ms{{87.93 }}{0.95} & \ms{{88.92 }}{1.12} \\ 

& \cmark& \cmark& \cmark &  \ms{{ \bf 93.94 }}{0.06} & \ms{{\bf 93.02 }}{0.15} & \ms{{ \bf 92.71 }}{0.10} \\

\bottomrule

\end{tabular}
}
\label{table:main_component_effectiveness}
\end{table}

\stitle{Ablation study.} We evaluate the contributions of the three main components introduced in Section~\ref{sec:framework_introduction}. Table~\ref{table:main_component_effectiveness} presents the results of an ablation study that assesses the individual and combined impacts of each component. When component 1 is excluded, we train with all labeled nodes without filtering. The results confirm the importance of all three components in enhancing the model's capacity to address class imbalance in graphs. Each component individually improves performance, with component 3 (synthetic nodes) generally providing the most significant enhancement. This improvement is likely due to the introduction of diverse patterns into minority classes during node synthesis, which helps balance the class distribution. Furthermore, the combination of components 2 and 3 nearly matches the performance of the full model. Adding component 1 to this combination offers a modest incremental benefit, usually within a one-percent range.



\begin{figure}[!ht]
  \centering
  \includegraphics[width=\linewidth]{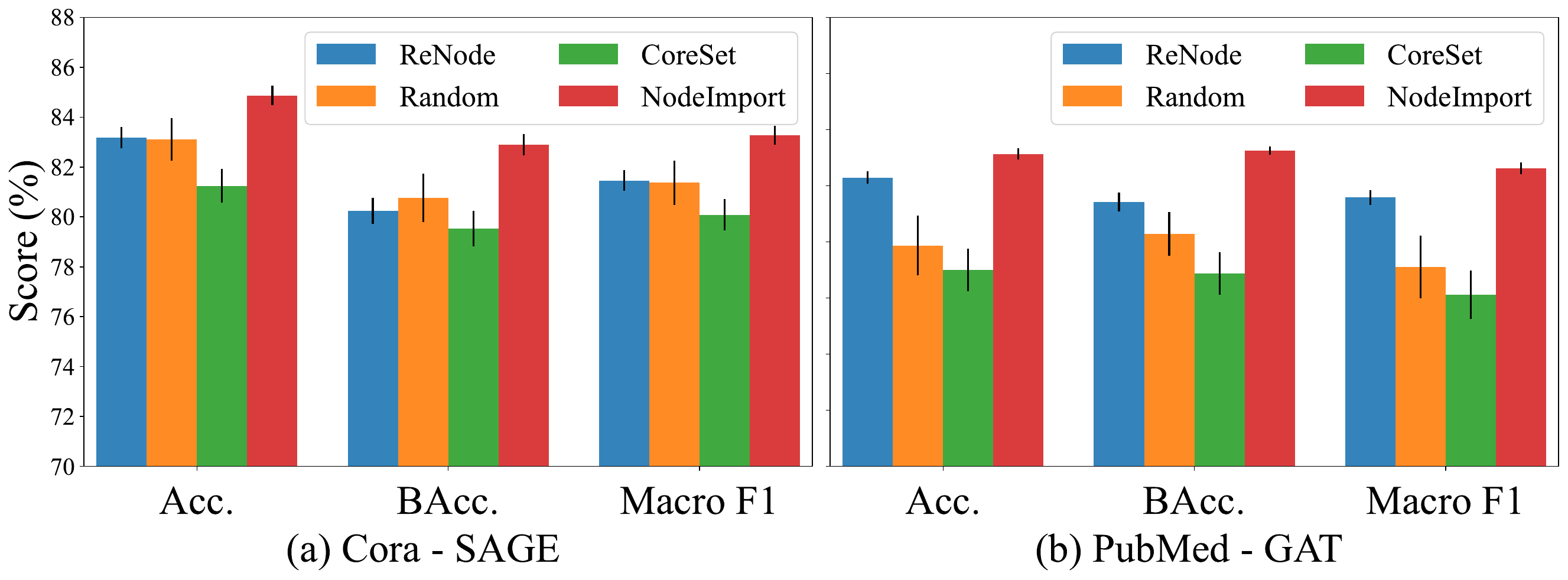}
  \caption{Comparison of meta-set construction methods.}
  \Description{Meta-set construction method comparison.}
  \label{fig.ablation_2}
\end{figure}

\stitle{Analysis of the meta-set construction method.} 
We evaluate the effectiveness of our meta-set construction method proposed in Section~\ref{sec:meta-set-construction} by comparing it with three alternatives: Random sampling, ReNode~\citep{chen2021topology} (which uses Totoro metrics to select top-ranked nodes), and CoreSet~\citep{sener2017active} (which iteratively selects nodes farthest from current selections). Results in Figure~\ref{fig.ablation_2} show that our proposed method consistently outperforms the other techniques across all metrics. Notably, Random sampling and ReNode exhibit similar performance levels, while CoreSet performs the worst. The inferior performance of CoreSet is likely due to the inclusion of many outliers in the meta-set, which negatively impacts the performance.


\begin{table}[!ht]
\center
\caption{F1 scores for each class on the Cora dataset using the GraphSAGE backbone, under an imbalance ratio of 50. The table header presents each class, labeled from $C_0$ to $C_6$, with its corresponding distribution located below each class label.}
\setlength{\tabcolsep}{2pt}
\resizebox{\linewidth}{!}{
\begin{tabular}{lccccccc}
\toprule
\textbf{Class} & $C_0$ & $C_1$ & $C_2$ & $C_3$ & $C_4$ & $C_5$ & $C_6$ \\
\textbf{Distribution} & $6.80\%$ & $1.80\%$ & $25.24\%$ & $48.54\%$ & $13.18\%$ & $3.47\%$ & $0.97\%$ \\
\midrule

Vanilla & 67.14 & 69.64 & 92.21 & 81.36 & 83.47 & 78.95 & 53.58 \\

GRAND & 64.57 & 52.59 & 93.10 & 73.54 & 84.34 & 72.94 & 14.75 \\

\cmidrule{1-8}

ReWeight & 71.68 & 78.88 & 93.43 & 84.13 & 84.45 & 80.49 & 65.88 \\

PC Softmax & 72.63 & 80.58 & 93.51 & 86.53 & 84.54 & 79.77 & 65.03 \\

Balanced Softmax & 72.18 & 80.84 & 93.12 & 85.41 & 84.36 & 80.83 & 68.38 \\

TAM(BS) & 72.69 & 80.34 & 92.62 & 85.05 & 84.78 & 81.52 & 70.44 \\

ReNode & 71.96 & 77.94 & 93.34 & 83.95 & 84.52 & 80.50 & 67.03 \\

TAM(ReNode) & 73.62 & 78.35 & 92.75 & 83.68 & 85.01 & 81.40 & 70.80 \\

\cmidrule{1-8}

GraphENS & 72.00 & 81.14 & 93.38 & 85.37 & 83.98 & 79.07 & 68.18 \\

TAM(G-ENS) & 72.14 & 82.33 & {\bf 93.54} & 85.45 & 84.32 & 81.60 & 70.07 \\

GraphSHA & 71.28 & 80.09 & 93.19 & 85.43 & 84.73 & 79.91 & 66.00 \\

\cmidrule{1-8}

\model & {\bf 75.99} & {\bf 83.84} & {93.11} & {\bf 87.12} & {\bf 86.28} & {\bf 82.48} & {\bf 74.09} \\

\bottomrule
\end{tabular}
}
\label{table:main-case-study}
\end{table}

\stitle{Case study on per-class F1 scores.}
Table~\ref{table:main-case-study} presents the per-class F1 scores on the Cora dataset using the GraphSAGE architecture. Baseline methods designed to mitigate class imbalance demonstrate improvements in F1 scores for both minority and majority classes, indicating that addressing class imbalance can refine decision boundaries and benefit all classes. Our model not only achieves the most significant improvements across most classes but also substantially reduces the performance gap between minority and majority classes. Notably, the improvement from our method is particularly pronounced in smaller classes, such as $C_0$, $C_1$, and $C_6$. Compared to the Vanilla method, the GRAND method shows a sharp decrease in the F1 score for the smallest class, $C_6$. This suggests that directly using unlabeled nodes for model training without considering class imbalance can lead to the misclassification of unlabeled nodes from extremely small classes like $C_6$, further exacerbating the class imbalance issue.

\section{Conclusions}


In this work, we introduce {\model}, a framework for class-imbalanced node classification. Our method dynamically evaluates node importance and incorporates key nodes into the training process to address class imbalance. We develop an efficient formula for assessing node importance and a framework for selecting high-quality labeled, unlabeled, and synthetic nodes. Additionally, we extract representative meta-samples from the training set. Experiments on benchmark datasets demonstrate that {\model} outperforms state-of-the-art methods, effectively mitigating class imbalance in node classification tasks.

\begin{acks}
This research is supported by the National Research Foundation, Singapore and Infocomm Media Development Authority under its Trust Tech Funding Initiative, the National Research Foundation, Singapore under its AI Singapore Programme (AISG Award No: AISG2-TC-2021-002).
\end{acks}

\bibliographystyle{ACM-Reference-Format}
\bibliography{reference}

\appendix

\section{Proofs for Theoretical Analysis}
In this section, we present theoretical proofs for the equations used in the manuscript. 
\subsection{Gradient Derivation of the Cross-Entropy Loss Function} \label{app.derivative_derivation}
To derive Eq.~\ref{eq:expression_3}, we need to find the derivative of the cross-entropy loss function $L[\cdot]$, as defined in Eq.~\ref{eq:loss_format}, with respect to the model parameters $\theta$. More generally, when dealing with a subset of $p$ nodes represented by $\tilde{A} \in \mathbb{R}^{p\times n}$ and their associated labels $Y \in [0,1]^{p\times c}$, the computation of the loss function can be broken down into several steps. At the first step, we calculate the output logits of the GNN encoder $S\in \mathbb{R}^{p\times c}$:
\begin{equation}
    S = \tilde{A} X \theta
\end{equation}
Next, we employ element-wise exponential operation to $S$ and obtain $E\in \mathbb{R}^{p\times c}$:
\begin{equation}
    E = \text{exp}(S)
\end{equation}
Then, we normalize $E$ along each row and get $H\in \mathbb{R}^{p\times c}$, which can be seen as the softmax of the output logits $S$:
\begin{equation}
    H = E \oslash EJ
\end{equation}
where $\oslash$ denotes the Hadamard division and $J\in\mathbb{R}^{c\times c}$ represents an all-one matrix. After that, we apply the element-wise logarithm to $H$ and get $Z \in \mathbb{R}^{p\times c}$
\begin{equation}
    Z = \text{log}(H)
\end{equation}
Finally, the cross-entropy loss $L$ is calculated as follows:
\begin{equation} \label{eq:loss_formula}
    L = - \langle Y, Z\rangle_F
\end{equation}
where $\langle  \cdot,\cdot \rangle _F$ denotes the Frobenius inner product. For simplicity, the symbol $:$ can be used to represent the Frobenius inner product $\langle  \cdot,\cdot \rangle _F$. Henceforth, these two symbols will be used interchangeably to indicate the Frobenius inner product.

Since directly computing the derivatives $\nabla$ is intractable and unrealistic, we choose to work in the differential form \footnote{We draw the inspiration from this blog \url{https://math.stackexchange.com/a/3850121/1237687}}. In particular, starting from the cross-entropy loss (Eq.~\ref{eq:loss_formula}), we have:
\begin{equation} \label{eq:first_diff}
    dL = -Y : dZ
\end{equation}
Next, 
\begin{equation}
    dZ = dH \oslash H
\end{equation}
Then, 
\begin{equation}
\begin{aligned}
    dH &= dE \odot EJ \oslash EJ \oslash EJ - d(EJ) \odot E \oslash EJ \oslash EJ \\
    &= dE \oslash EJ - (dE)J \odot H \oslash EJ \\ 
\end{aligned}
\end{equation}
where $\odot$ represents the Hadamard product. After that,
\begin{equation}
\begin{aligned}
    dE &= E \odot dS 
\end{aligned}
\end{equation}
Finally, 
\begin{equation} \label{eq:last_diff}
\begin{aligned}
    dS &= \tilde{A} X d\theta 
\end{aligned}
\end{equation}
Putting Eq.~\ref{eq:first_diff} to Eq.~\ref{eq:last_diff} together, we have the following:
\begin{equation}
\begin{aligned}
    dL &= -Y : dZ \\
    &= -Y : dH \oslash H \\
    &= -Y : (dE \oslash EJ - (dE)J \odot H \oslash EJ) \oslash H \\
    &= -Y : ((E \odot dS) \oslash EJ - (E \odot dS)J \odot H \oslash EJ) \oslash H \\
    &= -Y : (E \odot dS) \oslash EJ \oslash H - (E \odot dS)J \oslash EJ) \\
    &= -Y : dS - (E \odot dS)J \oslash EJ) \\
    &= -Y : dS + Y  : (E \odot dS)J \oslash EJ \\
    &= -Y : dS + (Y \oslash EJ)J \odot E : (dS) \\ 
    &= ((Y \oslash EJ)J \odot E -Y) : (dS) \\ 
    &= ((Y \oslash EJ)J \odot E -Y) : \tilde{A} X d\theta \\ 
    &= (\tilde{A} X)^T((Y \oslash EJ)J \odot E -Y) : d\theta \\ 
\end{aligned}
\end{equation}
In this way, we can get the gradient $\nabla L$ of the loss function in terms of the denominator layout:
\begin{equation} \label{eq:matrix_gradient_1}
\begin{aligned}
    \nabla_\theta L&= \frac{dL}{d\theta} \\ 
    &= (\tilde{A} X)^T((Y \oslash EJ)J \odot E -Y)
\end{aligned}
\end{equation}
Given that $Y$ is the label matrix with each row as a one-hot vector, we can simplify Eq.\ref{eq:matrix_gradient_1} and get:
\begin{equation} \label{eq:matrix_gradient_2}
\begin{aligned}
    \nabla_\theta L &= (\tilde{A} X)^T(H-Y) \\ 
\end{aligned}
\end{equation}
Hence, the proof is complete.

\section{Algorithm} \label{desc-algorithm}

\stitle{Pseudo-Code}
Algorithm~\ref{alg:main_algorithm} describes our approach. First, in lines 1-8, we derive the aggregation matrix and apply the PAM algorithm to each class's context embedding space to create a balanced meta-set. In lines 10-19, we generate synthetic nodes using a MixUp-like method. Subsequently, in lines 20-21, we assign pseudo-labels to the unlabeled nodes based on the highest predicted probabilities. In lines 22-24, we compute the importance scores of the nodes and filter out the important ones according to these scores. Lastly, in lines 25-29, we apply the cross-entropy loss to the filtered nodes and update the GNN model accordingly.

\begin{algorithm*}[th]
\caption{Pseudo-code of the framework}
\label{alg:main_algorithm}
\begin{algorithmic}[1]

\REQUIRE Input graph $\mathcal{G}=(\mathcal{V}, \mathcal{E}, A, X)$, class set $\mathcal{C}=\{1,\ldots, c\}$, training set $\gD_{tr}=\{(v,y_v)\}$, set of unlabeled nodes $\gD_{ul}$, GNN model $g(\cdot, \theta)$, depth of the aggregation matrix $K$, teleport probability when calculating the aggregation matrix $\alpha$, meta-set size per class $\tau$, scale for loss on unlabeled nodes $\beta$, scale for loss on synthetic nodes $\gamma$, learning rate $\kappa$, vector with all ones $J$, degree matrix $D$

\ENSURE Trained GNN parameters $\theta$

\STATE{Calculate the aggregation matrix $\tilde{A}=\frac{1}{K} \sum_{k=1}^{K} ((1-\alpha) ({D}^{-\frac{1}{2}}{A}{D}^{-\frac{1}{2}})^k + \alpha I)$}

\STATE{Initialize the set of meta-samples $\gM \leftarrow \emptyset$}

\FOR{$c \in \mathcal{C}$}

\STATE Apply PAM algorithm to find meta-samples in the class $\tilde{\gM}_k \leftarrow \text{PAM}(\tilde{A}X, \mathcal{V}^k, \tau)$

\STATE $\mathcal{M} \leftarrow \mathcal{M} \cup \tilde{\gM}^{k}$
\ENDFOR

\STATE{Construct the meta-set $\gD_{meta}$ from $\gM$.}

\STATE{Remove meta-samples from the training set $\gD_l=\gD_{tr} \setminus \gD_{meta}$}

\WHILE{not converged}
\STATE Sample a set of node pairs $\gS=\{\langle(v_s,y_s), (v_t, y_t)\rangle\}$ from the training set $D_{tr}$ based on Eq.\ref{eq:sample_probability}

\STATE Initialize the set of synthetic nodes $\gD_{syn}\leftarrow \emptyset$

\FOR{$\langle (v_s, y_s), (v_t, y_t) \rangle \in \gS$}

\STATE Sample the mixing ratio $\lambda \sim \text{Beta}(2,2)$xw

\STATE Generate synthetic node feature $x_{syn} = \lambda \cdot x_{s} + (1-\lambda) \cdot x_{t}$

\STATE Construct distribution for neighbor sampling $A'_{syn} = \lambda \cdot A_{s} + (1-\lambda) \cdot A_{t}$

\STATE Generate synthetic node label $y_{syn} = \lambda \cdot y_{s} + (1-\lambda) \cdot y_{t}$

\STATE Generate synthetic edges by sampling neighbors from the calculated distribution $A_{syn} = \text{Sample}(A'_{syn})$ 

\STATE $\gD_{syn} \leftarrow \gD_{syn} \cup \{(v_{syn}, y_{syn})\}$
\ENDFOR

\STATE Obtain predictions from the GNN model $H\leftarrow g(\gG, \theta)$

\STATE Generate pseudo-labels for unlabeled nodes $\gD_{ul}$ according to Eq.\ref{eq:pesudo-label}

\STATE Construct the set of valuable labeled nodes $\tilde{D}_{l} = \{(v, y_v) \in D_{l}\ |\ \eta_v >0 \}$

\STATE Construct the set of valuable unlabeled nodes $\tilde{D}_{ul} = \{(v, \hat{y}_{v})\ |\ v \in D_{ul} \text{ and } \eta_v >0 \}$

\STATE Construct the set of valuable synthetic nodes $\tilde{D}_{syn} = \{(v_{syn}, {y}_{syn})\ \in \gD_{syn}\ |\ \eta_{v_{syn}} >0 \}$ 

\STATE Compute loss for labeled nodes $\gL_{l}\leftarrow \sum_{(v,y_v)\in\tilde{\gD}_l} \text{Cross-Entropy}(y_v, h_v)$

\STATE Compute loss for unlabeled nodes $\mathcal{L}_{ul}\leftarrow \sum_{(v,\hat{y}_v)\in\tilde{D}_{ul}} \text{Cross-Entropy}(\hat{y}_v, h_{u})$

\STATE Compute loss for synthetic nodes $\mathcal{L}_{syn}\leftarrow \sum_{(v_{syn},y_{syn})\in\tilde{\gD}_{syn}} \text{Cross-Entropy}(y_{syn}, h_{v_{syn}})$

\STATE Compute the final loss $\mathcal{L}\leftarrow\mathcal{L}_{l}+\beta\cdot \mathcal{L}_{ul}+\gamma\cdot\mathcal{L}_{syn}$

\STATE Update the GNN model $\theta\leftarrow\theta-\kappa\cdot \nabla_{\theta}\mathcal{L}$
\ENDWHILE

\RETURN $\theta$

\end{algorithmic}
\end{algorithm*}

\stitle{Complexity analysis} 
Compared to the backbone GNN model, our algorithm adds additional computations in three main areas: meta-set construction (lines 1-8), synthetic node generation (lines 10-19), and node importance computation (lines 22-24). We will examine each part individually and elaborate on their impact on computational overhead. Let $\gV_{tr}$ represent the set of nodes within the training set $D_{tr}$, and $\gV_{ul}$ denote the set of unlabeled nodes.

\begin{itemize}[left=0.1pt]
    \item \textbf{Meta-set construction.} The adjacency matrix $A$ is sparse with $|\mathcal{E}|$ non-zero elements, the degree matrix $D$ is diagonal, and the feature matrix $X$ is dense. Thus, computing the context embedding $\tilde{A}X$ involves sparse-dense matrix multiplications, with a complexity of $\mathcal{O}(K|\mathcal{E}|d)$ where $K$ is the number of steps and $d$ is the feature dimension. Running the PAM algorithm on a class of size $|\mathcal{V}^{k}|$ for $\tau$ meta-samples incurs a complexity of $\mathcal{O}(\tau^2|\mathcal{V}^{k}|^2)$. Therefore, the overall complexity of the meta-set construction procedure is $\mathcal{O}(K|\mathcal{E}|d+\sum_{c\in\mathcal{C}}\tau|\mathcal{V}_c^{L}|^2)$.
    
    \item \textbf{Synthetic node generation.} The number of synthetic nodes generated per class is $\mathcal{O}(|\mathcal{V}_{tr}|)$. The process begins with sampling pairs of nodes, which incurs a complexity of $\mathcal{O}(|\mathcal{V}_{tr}|^2)$. For each sampled pair, generating synthetic node features and labels involves simple arithmetic operations with a complexity of $\mathcal{O}(|\mathcal{V}_{tr}|d)$. Furthermore, generating edges for these synthetic nodes to maintain the original degree distribution of the graph has a complexity of $\mathcal{O}(|\mathcal{V}_{tr}| \cdot \bar{d})$, where $\bar{d}$ represents the average degree of the graph. Therefore, the overall complexity of producing synthetic nodes is $\mathcal{O}(|\mathcal{V}_{tr}|^2+|\mathcal{V}_{tr}|d+|\mathcal{V}_{tr}| \cdot \bar{d})$.
    \item \textbf{Node importance computation.} Given that the amount of synthetic nodes is $\mathcal{O}(|\mathcal{V}_{tr}|)$ and these synthetic edges maintain the original degree distribution, the augmented graph with synthetic nodes consists of $\mathcal{O}(|\mathcal{E}|)$ edges. Moreover, generating pseudo-labels for unlabeled nodes requires a complexity of $\mathcal{O}(|\mathcal{V}_{ul}|c)$. Calculating node importance scores requires computing the context embedding for the augmented graph, which incurs a complexity of $\mathcal{O}(K|\mathcal{E}|d)$, and performing dense matrix multiplications and element-wise operations, which incurs a complexity of $\mathcal{O}(|\mathcal{V}|d\cdot c\tau+|\mathcal{V}|c^2\tau)$. Consequently, the complexity to determine node importance is $\mathcal{O}(K|\mathcal{E}|d+|\mathcal{V}_{ul}|c+|\mathcal{V}|d\cdot c\tau+|\mathcal{V}|c^2\tau)$.

\end{itemize}
Given that $|\mathcal{E}|\gg|\mathcal{V}_{tr}|$ and $c$, $\tau$, and $d$ are typically small constants, in a class-imbalanced node classification scenario, the added overhead for $T$ training epochs is simplified to $\mathcal{O}(T \cdot K|\mathcal{E}|d)$, which scales linearly with the number of graph edges. Therefore, the computational overhead is comparable to that of the primary GNN model, resulting in a lightweight additional computation.

\section{Descriptions of Datasets} \label{desc-datasets}

Our experiment adopts five commonly used benchmark datasets: three citation graphs and two Amazon co-purchase graphs. The dataset statistics are listed in Table~\ref{table:data_statistics}. In our implementation, we load these datasets via the PyTorch Geometric library~\cite{fey2019fast}.

\begin{itemize}[left=0.1pt]
\item \textbf{Cora}, \textbf{CiteSeer}, and \textbf{PubMed}~\cite{yang2016revisiting}: These three citation datasets consist of nodes representing academic papers and edges indicating the citation links. In Cora and CiteSeer, node features are binary vectors indicating the presence of specific words, whereas in PubMed, they are represented by TF/IDF weighted word vectors. The objective of these datasets is to classify papers into their respective categories. These datasets are publicly accessible at \url{https://github.com/kimiyoung/planetoid/raw/master/data}. For each citation dataset, we use the ten public splits created by ~\cite{pei2020geom}.

\item \textbf{Amazon-Photo} and \textbf{Amazon-Computer}~\cite{shchur2018pitfalls}: These two co-purchase datasets consist of nodes representing goods and edges representing frequent co-purchase patterns between pairs of products. Node features in these datasets are encoded as bag-of-words representations derived from corresponding product reviews. The objective of these datasets is to classify products into their respective categories. These datasets are publicly accessible at \url{https://github.com/shchur/gnn-benchmark/raw/master/data}.  

\end{itemize}

\section{Descriptions of the Baselines} \label{desc-baselines}

This section will introduce baselines used in more detail and their hyper-parameter configurations.

\begin{itemize}[left=0.1pt]
    \item \textbf{Graph Random Neural Networks (GRAND)~\cite{feng2020grand}}: This approach employs consistency regularization to ensure consistent model predictions across various augmented versions of an unlabeled node. These versions are created through data augmentation, and the average of their prediction outcomes is used as the pseudo-label for the unlabeled node. In the experiments, the sharpening temperature is set to 0.5, the scale of the unlabeled loss is set to 1, and the augmentation sampling for each unlabeled node occurs 4 times.
    \item \textbf{Re-weight~\cite{japkowicz2002class}}: In this approach, higher weights are assigned to samples from minority classes. During the experiment, the weight of each sample is inversely proportional to the number of training samples in its class.
    \item \textbf{Balanced Softmax~\cite{ren2020balanced}}: This approach seeks to reduce the distribution shift between the training and test data. It adjusts the output logits by adding the logarithm of the class size before applying Softmax to calculate the cross-entropy loss. 
    \item \textbf{Post-Compensated Softmax (PC Softmax)~\cite{hong2021disentangling}}: This approach disentangles the training label distribution from model predictions and incorporates the testing label distribution into the disentangled predictions. It adjusts the output logits during the inference stage by subtracting the logarithm of the class size before applying Softmax.
    \item \textbf{ReNode~\cite{chen2021topology}}: This work devises a re-weighting strategy based on nodes' relative topological positions within their classes. Nodes closer to the topological class centers tend to receive higher weights, while those near class boundaries will receive lower weights. In the experiment, the cosine annealing lower bound $w_{min}$ is set to 0.5, and the upper bound $w_{max}$ is set to 1.5. The teleport probability $\alpha$ in the PageRank is set to 0.15.
    \item \textbf{Topology-Aware Margin (TAM)~\cite{song2022tam}}: This approach adjusts the logits of training nodes when computing the loss function using two types of margins. The Anomalous Connectivity Margin (ACM) accounts for a node's neighbor label distribution and the average connectivity patterns within its class. The Anomalous Distribution-aware Margin (ADM) considers a node's relative distance to other classes compared to its own class based on neighbor label distribution. In the experiment, we integrate TAM with Balanced Softmax, ReNode, and GraphENS for a comprehensive evaluation. We set the coefficient of the ACM term $\alpha$ to 2.5, the coefficient of the ADM term $\beta$ to 0.5, the minimum temperature of the class-wise temperature $\phi$ to 1.2, and the warmup period to 5 iterations.
    \item \Nan{\textbf{Regularize Variance (ReVar)~\citep{yan2023rethinking}}: This work leverages data augmentation to estimate model variance and uses the estimated variance as a regularization term during training. The two intensity terms $\lambda_1$ and $\lambda_2$ are set to $0.25$ and $2.85$, respectively.}
    \item \Nan{\textbf{BAlanced Topological augmentation (BAT)~\citep{Liu24BAT}}: This work addresses class imbalance by identifying and correcting nodes with ineffective message passing due to topological disparities. We adapt BAT with 1st-order estimation during experiment.}
    \item \textbf{GraphENS~\cite{park2022graphens}}: This over-sampling method generates new minority nodes by combining nodes from minority classes with other nodes based on their similarity, preserving the semantics of the minority classes. It also uses a gradient-based approach to determine the importance of each feature dimension to model prediction and randomly masks important feature dimensions during feature mixing. In the experiment, the mixing ratio $\lambda$ is sampled from a $\text{Beta}(2,2)$ distribution, the feature masking hyperparameter $K$ is set to 1, the prediction temperature $\tau$ is set to 2, and the warmup period is set to 5 iterations.
    \item \textbf{GraphSHA~\cite{li2023graphsha}}: This method synthesizes new nodes by combining minority nodes (anchors) with other nodes (auxiliaries). The sampling of node pairs is based on the predictive output from the GNN encoder. Each synthetic node's features are generated through a convex combination of features. Edges are sampled only from the anchor node to construct the new neighborhood, utilizing a smoothed version of the adjacency matrix via graph diffusion-based smoothing. In the experiment, the mixing ratio $\delta$ is sampled from a $\text{Beta}(1,100)$ distribution. The Personalized PageRank (PPR) version of the adjacency matrix is used, and the temperature $T$ of the SoftMax function is set to 2 for the calculating model confidence.
\end{itemize}

\section{Deatials on the Evaluation Process} \label{desc-evaluation}

\subsection{Experimental Environment}
We conducted all experiments on a server running Ubuntu 22.04.3, equipped with a 2.90GHz Intel Xeon Gold 6226R CPU, 512GB of RAM, and 8 NVIDIA GeForce RTX 3090 GPUs, each with 24GB of memory. Our model is implemented using Python 3.8.0, PyTorch 1.13.0 with CUDA 11.7, and PyTorch Geometric 2.4.0. Additional libraries used include scikit-learn and scikit-learn-extra.

\subsection{Configurations of GNN Backbones}
\begin{itemize}[left=0.1pt]
    \item {\bf Graph Convolutional Network (GCN)~\cite{kipf2017semisupervised}:} The network is built with two GCN layers, with a hidden dimension of 256. Following the first convolutional layer, a ReLU activation function is used, coupled with a dropout layer with a dropping rate of 0.5. 
    \item {\bf Graph Attention Networks (GAT)~\cite{velivckovic2017graph}:} The network is constructed with two GAT layers, with a hidden dimension size of 256. The multi-head attention mechanism is applied, with the number of heads set to 4. Following the first convolutional layer, a ReLU activation function is used, coupled with a dropout layer with a dropping rate of 0.5. 
    \item {\bf Graph Sample and AggregatE (GraphSAGE)~\cite{hamilton2017inductive}):} The network is structured with two SAGE layers, with a hidden dimension size of 256. The mean aggregator is used to aggregate neighboring features. Following the first convolutional layer, a ReLU activation function is utilized, coupled with a dropout layer with a dropping rate of 0.5. 
\end{itemize}

\subsection{Configuration of Our Model}
For the computation of the aggregation matrix $\tilde{A}$, the depth $K$ is selected from $\{2,4,8,16\}$m and the teleport probability $\alpha$ is selected from $\{0.05, 0.10, 0.15, 0.20\}$. The scaling factors for the losses of the unlabeled and synthetic nodes, $\beta$ and $\gamma$, are selected from $\{0.5, 1.0, 2.0, 4.0\}$. \Nan{We control the number of meta-samples per class to around $40\%$ of the smallest class in the training set.}

\subsection{Training Details}
All methods are trained for 2000 epochs, and the parameters of GNN backbones are optimized via the Adam optimizer~\cite{Kingma2015Adam}. The initial learning rate is set to 0.01, subject to a halving adjustment upon observing no improvement over 100 epochs in terms of the validation loss. Except for parameters in the final graph convolutional layer, a weight decay of 0.0005 is applied to all learnable parameters. The average of accuracy and macro F1 score on the validation set is employed to select the final model for testing.

\begin{figure}
  \centering
  \includegraphics[width=\linewidth]{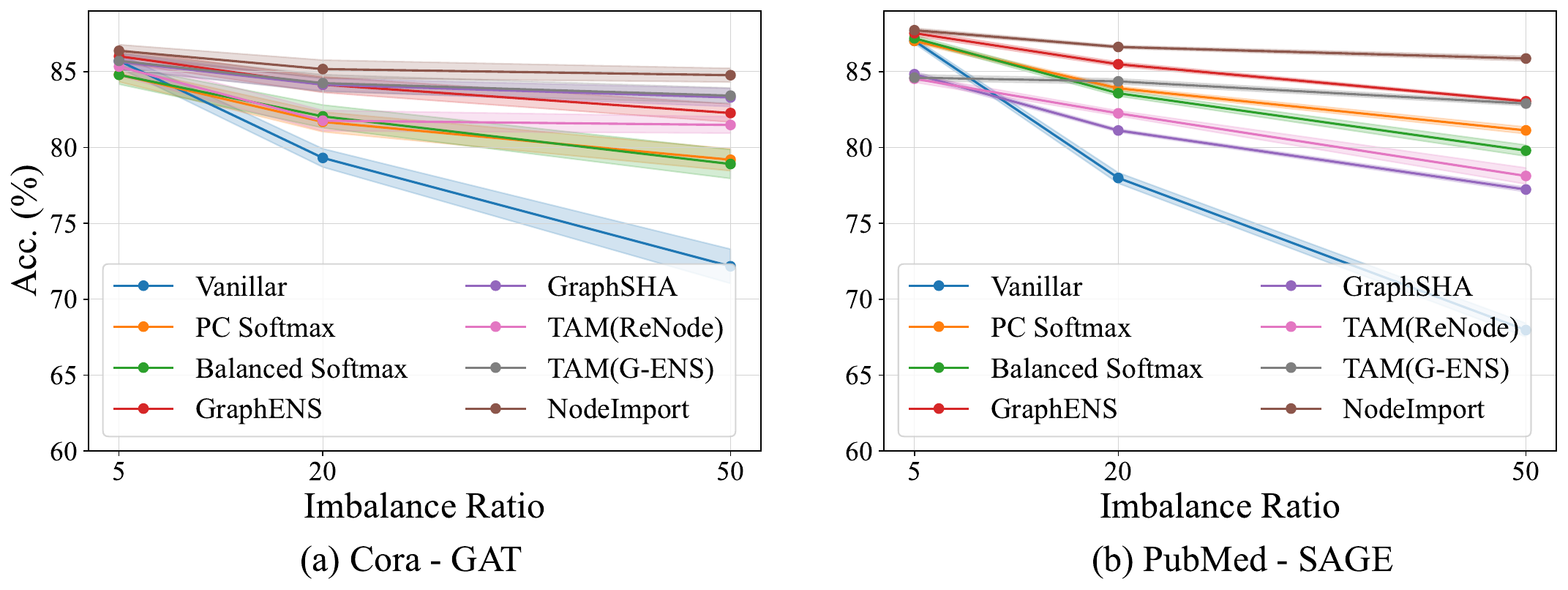}
  \caption{\Nan{Experiments under varying imbalance ratios in terms of accuracy.}}
  \Description{Acc results.}
  \label{figure:appendix-imb-ratio-acc}
\end{figure}

\begin{figure}
  \centering
  \includegraphics[width=\linewidth]{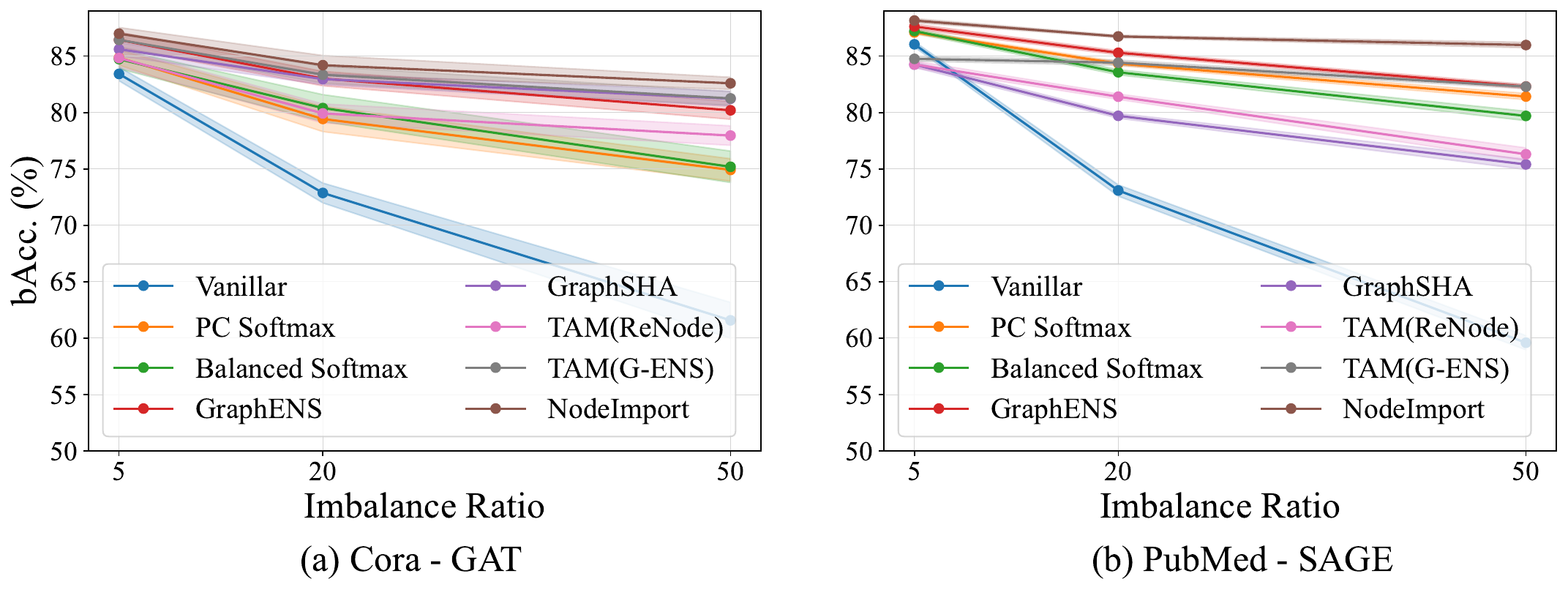}
  \caption{\Nan{Experiments under varying imbalance ratios in terms of balanced accuracy.}}
  \Description{Bacc results.}
  \label{figure:appendix-imb-ratio-bacc}
\end{figure}

\begin{table}[t]
  \caption{\Nan{Statistics of the ogbn-arxiv dataset}}
  \label{table:ogbn_data_statistics}
  \centering
  \begin{tabular}{crrrr}
  \toprule
{Dataset}  & {Nodes} & {Edges} & {Features} & {Classes} \\
\midrule
ogbn-arxiv & 169,343 & 1,166,243 & 128 & 40 \\
\bottomrule
\end{tabular}
\end{table}

\section{Additional Experimental Results} \label{app.add-experiments}

\subsection{\Nan{Experiments with a large-scale dataset}}
\label{app.large-scale-dataset}
\Nan{In this subsection, we evaluate our method on the large-scale arXiv dataset from the OGB benchmark~\citep{ogbnbenchmark}, summarized in Table~\ref{table:ogbn_data_statistics}. Due to its size, data-level methods like GraphENS and GraphSHA encounter memory limitations, and our full model faces similar issues with C3 due to node synthesis. As a result, we evaluated our model using only C1 and C2. Table~\ref{table:ogbn_performance} presents the results in terms of balanced accuracy, execution time, and peak GPU memory usage. Despite the absence of C3, our model demonstrates competitive performance, with execution time and memory usage comparable to the vanilla method. The ogbn-arxiv dataset also poses unique challenges for class-imbalanced node classification, including an extremely long-tailed distribution and varying majority and minority class compositions across the train, validation, and test sets, highlighting its potential for further investigation.}

\begin{table}[!h]
  \caption{\Nan{Experimental results on the ogbn-arxiv dataset, showcasing model performance across balanced accuracy, execution time, and peak GPU memory usage (columns 2–4).}}
  \label{table:ogbn_performance}
  \centering
  \setlength{\columnsep}{1pt}%
  \resizebox{0.98\linewidth}{!}{
  \begin{tabular}{cccc}
  \toprule
{Methods}  & {bAcc.} & {Exec Time(s)} & {Peak GPU Mem (GB)} \\
\midrule
Vanilla & \ms{44.43}{0.13} & \ms{3325}{13} & 6.29 \\
\cmidrule{1-4}
ReWeight & \ms{56.52}{0.16} & \ms{3333}{6} & 6.29 \\
TAM(BS) & \ms{53.45}{0.23} & \ms{3436}{18} & 6.29 \\
\cmidrule{1-4}
ours(w/o C3) & \ms{55.59}{0.14} & \ms{3473}{13} & 6.72 \\
\bottomrule
\end{tabular}
}
\end{table}

\begin{table*}[t]
\caption{Performance comparison (\%) on benchmark graphs with an imbalance ratio of 50. The best results are in bold. Models with performances within the standard error are statistically comparable.}
\begin{center}
\setlength{\columnsep}{1pt}%
\resizebox{0.55\linewidth}{!}{
\begin{tabular}{@{\extracolsep{0pt}}rlccccc@{}}
\toprule
 & \multirow{1}{*}{\textbf{Dataset}} & \multicolumn{1}{c}{Cora} & \multicolumn{1}{c}{CiteSeer} & \multicolumn{1}{c}{PubMed} & \multicolumn{1}{c}{Photo} & \multicolumn{1}{c}{Computers} \\ 
 
& \textbf{IR=50} & Acc. & Acc. & Acc. & Acc. & Acc. \\
\midrule

\multirow{14}{*}{\rotatebox[origin=c]{90}{GCN}} 

& Vanilla & \ms{{81.55 }}{0.51} & \ms{59.74}{1.32} & \ms{60.68}{0.51} & \ms{87.35}{0.12} & \ms{82.23}{0.26} \\

& GRAND & \ms{79.98}{0.58} & \ms{55.59}{1.35} & \ms{55.06}{0.93} & \ms{88.06}{0.07} & \ms{83.78}{0.17} \\

\cmidrule{2-7}

& ReWeight & \ms{83.74}{0.48} & \ms{67.33}{1.00} & \ms{81.93}{0.27} & \ms{91.71}{0.09} & \ms{85.24}{0.19} \\

& PC Softmax & \ms{81.99}{0.59} & \ms{65.39}{1.13} & \ms{77.40}{0.58} & \ms{90.04}{0.16} & \ms{85.27}{0.46} \\

& Balanced Softmax & \ms{83.72}{0.54} & \ms{69.09}{0.86} & \ms{82.53}{0.19} & \ms{90.55}{0.17} & \ms{84.65}{0.50} \\

& TAM(BS) & \ms{84.10}{0.51} & \ms{69.30}{0.77} & \ms{81.44}{0.31} & \ms{84.01}{0.12} & \ms{77.56}{0.44} \\

& ReNode & \ms{83.86}{0.38} & \ms{67.51}{1.05} & \ms{81.59}{0.37} & \ms{91.53}{0.04} & \ms{85.67}{0.18} \\

& TAM(ReNode) & \ms{83.90}{0.57} & \ms{68.00}{1.01} & \ms{81.79}{0.61} & \ms{90.00}{0.08} & \ms{82.19}{0.16} \\

& \Nan{ReVar}  & \Nan{ \ms{81.19}{0.57}}  & \Nan{ \ms{68.81}{0.93}}  & \Nan{ \ms{72.35}{0.74}}  & \Nan{ \ms{89.41}{0.05}}  & \Nan{ \ms{88.65}{0.08}}  \\

& \Nan{BAT}  & \Nan{ \ms{82.33}{0.31}}  & \Nan{ \ms{62.14}{1.44}}  & \Nan{ \ms{61.19}{0.68}}  & \Nan{ \ms{87.57}{0.11}}  & \Nan{ \ms{81.97}{0.34}}  \\

\cmidrule{2-7}

& GraphENS & \ms{83.34}{0.40} & \ms{69.43}{0.96} & \ms{82.24}{0.23} & \ms{91.84}{0.07} & \ms{84.84}{0.13} \\

& TAM(G-ENS) & \ms{83.66}{0.41} & \ms{69.84}{0.90} & \ms{82.95}{0.24} & \ms{87.45}{0.15} & \ms{81.36}{0.19} \\

& GraphSHA & \ms{{\bf 84.87}}{0.47} & \ms{67.92}{1.13} & \ms{79.86}{0.30} & \ms{90.77}{0.12} & \ms{85.72}{0.11} \\

\cmidrule{2-7}

& \model & \ms{{ \bf 85.11 }}{0.42} & \ms{{ \bf 71.47 }}{0.72} & \ms{{ \bf 84.42 }}{0.16} & \ms{{ \bf 92.71 }}{0.04} & \ms{{ \bf 88.80 }}{0.04} \\

\cline{1-7}
\noalign{\vskip\doublerulesep
         \vskip-\arrayrulewidth} \cline{1-7}
\rule{0pt}{2.5ex}  

\multirow{14}{*}{\rotatebox[origin=c]{90}{GAT}} 

& Vanilla & \ms{72.17}{1.13} & \ms{56.81}{1.36} & \ms{54.20}{1.74} & \ms{82.81}{0.57} & \ms{79.94}{0.29} \\

& GRAND & \ms{68.43}{0.95} & \ms{50.86}{1.62} & \ms{52.60}{1.56} & \ms{84.02}{0.48} & \ms{80.76}{0.38} \\

\cmidrule{2-7}

& ReWeight & \ms{83.14}{0.45} & \ms{66.98}{1.12} & \ms{80.98}{0.19} & \ms{86.38}{0.30} & \ms{81.92}{0.35} \\

& PC Softmax & \ms{79.20}{0.73} & \ms{66.87}{1.04} & \ms{80.24}{0.24} & \ms{84.73}{0.47} & \ms{78.95}{0.38} \\

& Balanced Softmax & \ms{78.91}{0.96} & \ms{66.76}{0.83} & \ms{81.01}{0.46} & \ms{87.01}{0.42} & \ms{81.63}{0.32} \\

& TAM(BS) & \ms{77.83}{0.73} & \ms{68.03}{0.92} & \ms{81.37}{0.36} & \ms{81.74}{0.42} & \ms{76.95}{0.86} \\

& ReNode & \ms{82.78}{0.59} & \ms{65.15}{1.12} & \ms{80.63}{0.22} & \ms{86.35}{0.50} & \ms{82.06}{0.29} \\

& TAM(ReNode) & \ms{81.49}{0.54} & \ms{65.02}{1.17} & \ms{80.65}{0.93} & \ms{82.81}{0.64} & \ms{79.71}{0.36} \\

& \Nan{ReVar}  & \Nan{ \ms{81.57}{0.39}}  & \Nan{ \ms{69.34}{0.98}}  & \Nan{ \ms{71.92}{1.10}}  & \Nan{ \ms{89.37}{0.02}}  & \Nan{ \ms{88.27}{0.11}}  \\

& \Nan{BAT}  & \Nan{ \ms{75.03}{1.21}}  & \Nan{ \ms{57.91}{1.53}}  & \Nan{ \ms{54.12}{1.78}}  & \Nan{ \ms{83.98}{0.62}}  & \Nan{ \ms{80.42}{0.47}}  \\

\cmidrule{2-7}

& GraphENS & \ms{82.27}{0.60} & \ms{68.25}{0.83} & \ms{81.78}{0.24} & \ms{90.68}{0.15} & \ms{86.01}{0.17} \\

& TAM(G-ENS) & \ms{83.42}{0.51} & \ms{68.57}{0.83} & \ms{82.56}{0.19} & \ms{87.47}{0.31} & \ms{82.86}{0.25} \\

& GraphSHA & \ms{83.30}{0.60} & \ms{66.52}{1.15} & \ms{81.76}{0.18} & \ms{89.22}{0.24} & \ms{85.06}{0.26} \\

\cmidrule{2-7}

& \model & \ms{{ \bf 84.21 }}{0.46} & \ms{{ \bf 71.54 }}{0.71} & \ms{{ \bf 83.92 }}{0.23} & \ms{{ \bf 92.76 }}{0.15} & \ms{{ \bf 89.55 }}{0.06} \\ 

\cline{1-7}
\noalign{\vskip\doublerulesep
         \vskip-\arrayrulewidth} \cline{1-7}
\rule{0pt}{2.5ex}  

\multirow{14}{*}{\rotatebox[origin=c]{90}{GraphSAGE}} 

& Vanilla & \ms{79.52}{0.44} & \ms{{56.07 }}{1.65} & \ms{67.97}{0.45} & \ms{{90.53 }}{0.30} & \ms{85.98}{0.17} \\

& GRAND & \ms{74.35}{0.72} & \ms{51.28}{1.33} & \ms{64.76}{0.34} & \ms{88.22}{0.36} & \ms{84.99}{0.16} \\

\cmidrule{2-7}

& ReWeight & \ms{82.29}{0.49} & \ms{62.10}{1.38} & \ms{78.94}{0.31} & \ms{92.77}{0.06} & \ms{88.00}{0.03} \\

& PC Softmax & \ms{82.78}{0.41} & \ms{64.11}{1.29} & \ms{81.13}{0.23} & \ms{93.15}{0.07} & \ms{87.46}{0.17} \\

& Balanced Softmax & \ms{82.74}{0.53} & \ms{64.63}{1.35} & \ms{79.80}{0.37} & \ms{92.91}{0.08} & \ms{86.87}{0.20} \\

& TAM(BS) & \ms{82.84}{0.37} & \ms{64.32}{1.41} & \ms{79.35}{0.38} & \ms{86.29}{0.19} & \ms{84.06}{0.33} \\

& ReNode & \ms{82.31}{0.43} & \ms{61.99}{1.68} & \ms{78.41}{0.15} & \ms{92.43}{0.07} & \ms{87.83}{0.04} \\

& TAM(ReNode) & \ms{82.72}{0.39} & \ms{62.66}{1.44} & \ms{78.13}{0.54} & \ms{90.54}{0.15} & \ms{84.83}{0.13} \\

& \Nan{ReVar}  & \Nan{ \ms{77.40}{0.55}}  & \Nan{ \ms{65.70}{1.03}}  & \Nan{ \ms{76.80}{0.65}}  & \Nan{ \ms{91.82}{0.02}}  & \Nan{ \ms{\bf 88.33}{0.13}}  \\

& \Nan{BAT}  & \Nan{ \ms{82.17}{0.52}}  & \Nan{ \ms{58.65}{1.61}}  & \Nan{ \ms{67.80}{0.49}}  & \Nan{ \ms{90.63}{0.09}}  & \Nan{ \ms{86.41}{0.18}}  \\

\cmidrule{2-7}

& GraphENS & \ms{82.49}{0.53} & \ms{68.89}{1.00} & \ms{83.05}{0.09} & \ms{93.29}{0.06} & \ms{87.47}{0.08} \\

& TAM(G-ENS) & \ms{83.06}{0.42} & \ms{69.67}{0.86} & \ms{82.90}{0.14} & \ms{90.49}{0.18} & \ms{85.05}{0.16} \\

& GraphSHA & \ms{82.43}{0.46} & \ms{62.95}{1.63} & \ms{77.23}{0.15} & \ms{93.11}{0.06} & \ms{{\bf 88.28}}{0.05} \\

\cmidrule{2-7}

& \model & \ms{{ \bf 84.87 }}{0.40} & \ms{{\bf 71.12 }}{0.84} & \ms{{ \bf 86.30 }}{0.13} & \ms{{ \bf 93.94 }}{0.06} & \ms{{ 88.00 }}{0.07} \\

\bottomrule
\end{tabular}
}
\end{center}
\label{table:main_imb_50_acc}
\vspace{-0.05in}
\end{table*}

\subsection{Overall Performance in Terms of Accuracy} \label{appendix-overall-acc}
Table~\ref{table:main_imb_50_acc} lists the model performance against selected baselines in terms of the accuracy metric. The result is consistent with that discussed in Section~\ref{sec:model_performance}.

\subsection{Performance Trends Under Different Imbalance Ratios} \label{appendix-performance-trend}

Figure~\ref{figure:appendix-imb-ratio-acc} and Figure~\ref{figure:appendix-imb-ratio-bacc} present the performance trends of different methods under various imbalance ratios in terms of the accuracy and balanced accuracy metrics, respectively. These findings align with the discussions presented in Section~\ref{sec:model_performance}.

\begin{figure}[!ht]
  \centering
  \includegraphics[width=\linewidth]{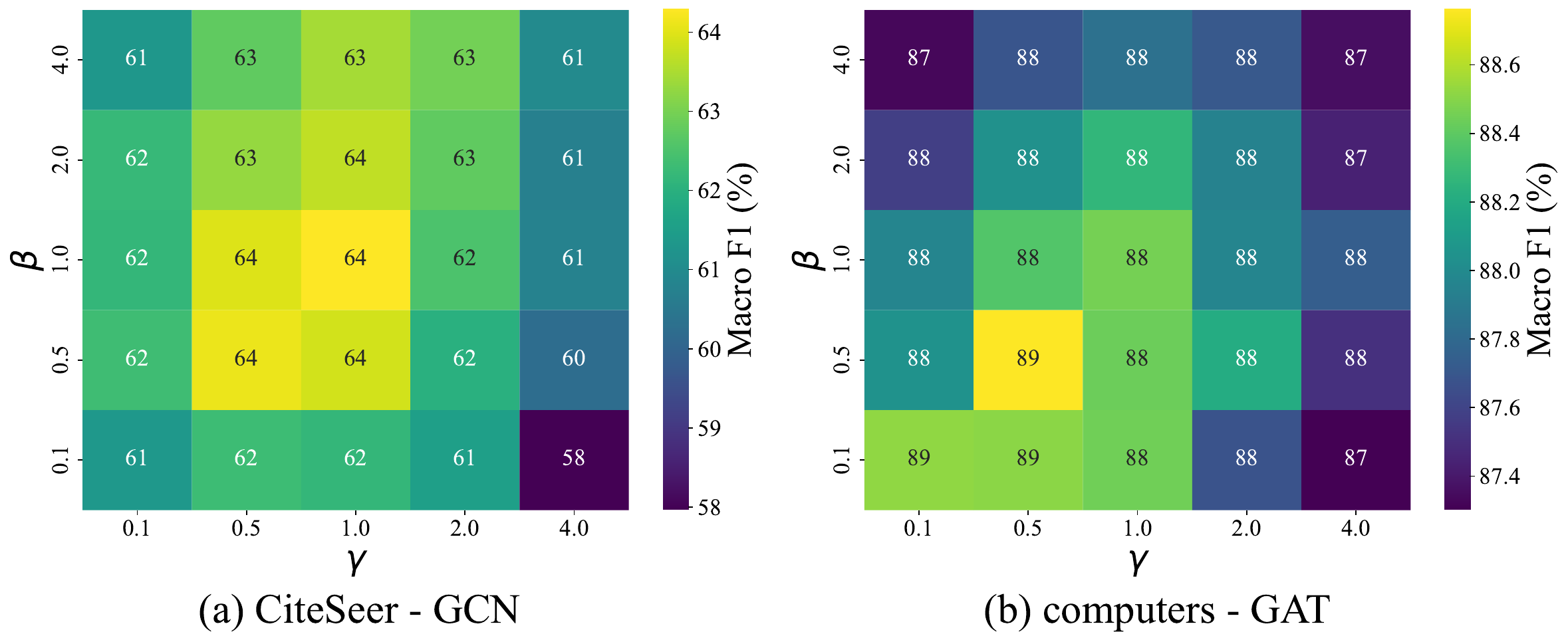}
  \caption{Sensitivity analysis of hyper-parameters $\beta$ and $\gamma$.}
  \Description{Hyper-parameter sensitivity.}
  \label{fig.hp_sen}
\end{figure}

\begin{figure}[!ht]
  \centering
  \includegraphics[width=\linewidth]{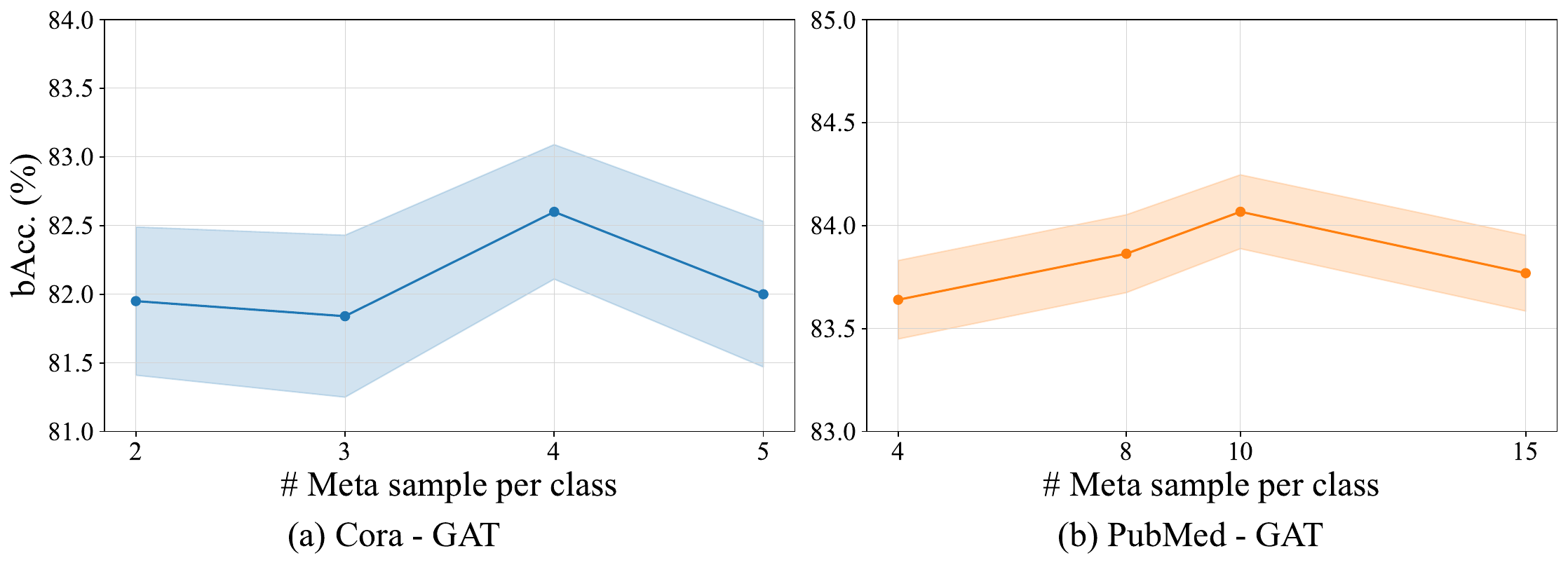}
  \caption{\Nan{Sensitivity analysis of the size of the meta-set.}}
  \Description{Hyper-parameter sensitivity.}
  \label{fig.hp_sen_meta}
\end{figure}

\subsection{Hyper-Parameters Sensitivity} \label{app.parameters-sensitivity}
This analysis examines how the two scaling factors $\beta$ and $\gamma$ of the loss function in Eq.~\ref{eq:final_loss} affect the model performance. Specifically, we explore a range of values for \(\beta\) and \(\gamma\): \{0.1, 0.5, 1.0, 2.0, 4.0\}, and assess the performance for each pairing of these factors. The results, as shown in Figure~\ref{fig.hp_sen}, illustrate that the model maintains stable performance when $\beta$ and $\gamma$ are set around $1.0$. However, as these parameters increase far beyond this point, a noticeable decrease in performance is observed. This decline is likely due to the model overfitting, where excessively high scaling factors disproportionately amplify the unlabeled or synthetic losses, thus degrading the capability of the model.

\Nan{Furthermore, we evaluate the impact of meta-set size on model performance, as shown in Figure~\ref{fig.hp_sen_meta}, which presents the results in terms of balanced accuracy. The findings indicate that moderate-sized meta-sets achieve better overall performance.}

\begin{figure}[!ht]
  \centering
  \includegraphics[width=\linewidth]{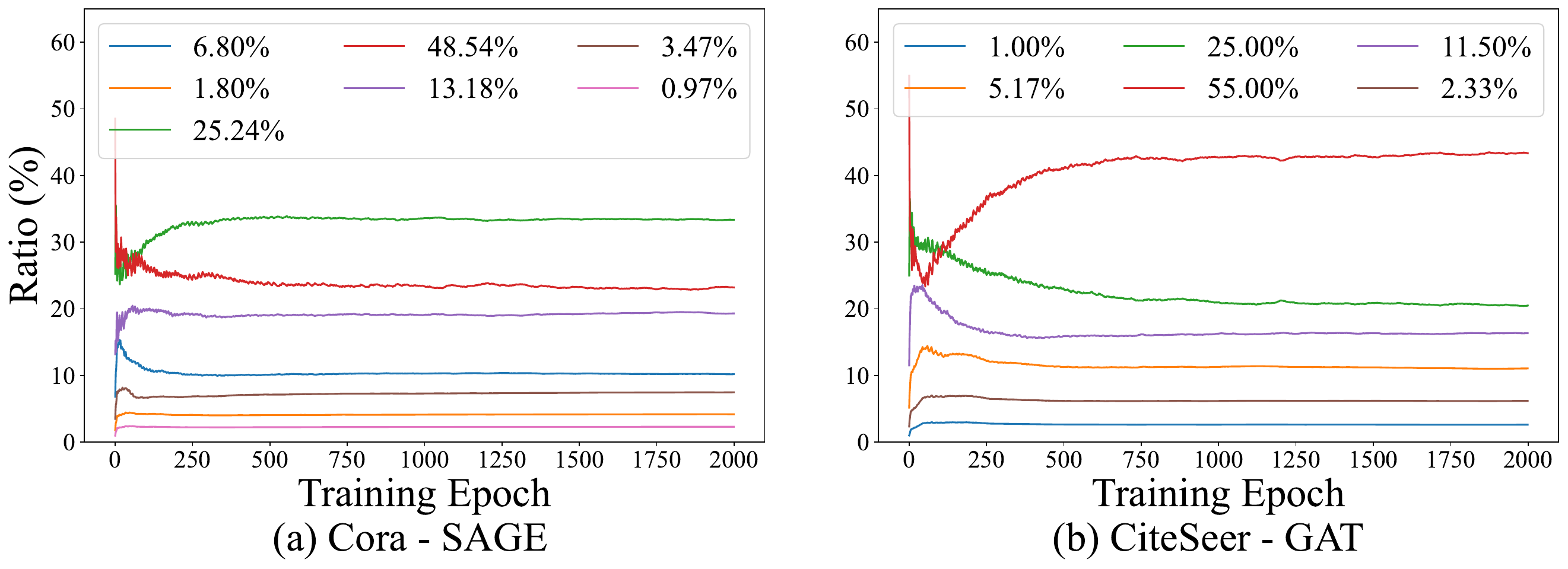}
  \caption{Cumulative class distribution of the filtered labeled nodes as training progresses. Each line represents the cumulative distribution of a class, with the percentage in the legend indicating its initial distribution in the training set.}
  \Description{Cumulative class distribution.}
  \label{figure:appendix-sample-rates}
\end{figure}

\subsection{Class Distribution of the Filtered Labeled Nodes} \label{app.class-distribution}
Figure~\ref{figure:appendix-sample-rates} shows the cumulative class distribution of the filtered labeled nodes over training epochs for the Cora (GraphSAGE) and CiteSeer (GAT) datasets. At each epoch, we count the number of nodes on which the model is trained for each class up to that point and normalize these counts among the classes to represent their relative proportions. In the Cora dataset (Figure~\ref{figure:appendix-sample-rates}a), the majority classes (\textcolor{red}{Red} and \textcolor{green}{Green} lines) are progressively down-sampled, as indicated by the decreasing trends. In contrast, minority classes (e.g., \textcolor{brown}{Brown} and \textcolor{blue}{Blue} lines) are up-sampled, showing increasing trends. Similarly, in the CiteSeer dataset (Figures~\ref{figure:appendix-sample-rates}b), the majority classes (\textcolor{red}{Red} and \textcolor{green}{Green} lines) are down-sampled, while minority classes (e.g., \textcolor{brown}{Brown} and \textcolor{orange}{Orange} lines) are up-sampled. In addition, the extent of up-sampling and down-sampling varies across classes, demonstrating our model's capability to dynamically adjust the training label distribution and effectively address node imbalance.

\end{document}